\documentclass[11pt]{article}

\usepackage[final]{acl}

\usepackage{times}
\usepackage{latexsym}

\usepackage[T1]{fontenc}
\usepackage[utf8]{inputenc}

\usepackage{microtype}

\usepackage{inconsolata}

\usepackage{graphicx}

\usepackage{titletoc}
\usepackage{mdframed}
\definecolor{acmblue}{RGB}{0,112,192}
\usepackage{booktabs} % For formal tables
\usepackage{multirow}
\usepackage{makecell}

\usepackage{enumitem}
\usepackage{url}
\usepackage{colortbl}      % 用于表格的颜色支持
\usepackage{graphicx}
\usepackage{booktabs} % 美化表格用
\usepackage{array} % 支持表格列宽调整
\renewcommand{\arraystretch}{1.1} % 增加单元格高度
\usepackage{caption}       % 控制标题格式
\usepackage{longtable} % 引入 longtable 包
\usepackage{pifont} % for \ding symbols
\usepackage[textsize=tiny]{todonotes}
\usepackage{amsthm} % 加在导言区

\newtheorem{remark}{Remark}
\usepackage{nameref} 

\usepackage{amsmath} 
\usepackage{mathtools}

\usepackage{titletoc}
\usepackage{mdframed}
\definecolor{acmblue}{RGB}{0,112,192}
\usepackage{booktabs} % For formal tables
\usepackage{multirow}
\usepackage{makecell}
\usepackage{amsmath}
\usepackage{amssymb}   % 或者 \usepackage{amsfonts}
\usepackage{enumitem}
\usepackage{url}
\usepackage{colortbl}      % 用于表格的颜色支持
\usepackage{graphicx}
\usepackage{booktabs} % 美化表格用
\usepackage{array} % 支持表格列宽调整
\renewcommand{\arraystretch}{1.1} % 增加单元格高度
\usepackage{caption}       % 控制标题格式
\usepackage{longtable} % 引入 longtable 包
\usepackage{pifont} % for \ding symbols
\usepackage[textsize=tiny]{todonotes}

\usepackage{listings}
\title{Exploring Graph Learning Tasks with Pure LLMs: A Comprehensive Benchmark and Investigation}

\author{
Yuxiang Wang$^{1}$, Xinnan Dai$^{2}$, Wenqi Fan$^{3}$, Yao Ma$^{4}$ \\
$^{1}$The Chinese University of Hong Kong, Shenzhen \\
$^{2}$Michigan State University, USA \\
$^{3}$The Hong Kong Polytechnic University, HK SAR \\
$^{4}$Rensselaer Polytechnic Institute, USA \\
\texttt{yuxiangwang1@link.cuhk.edu.cn, daixinna@msu.edu, wenqifan03@gmail.com, may13@rpi.edu}
}

\begin{document}
\maketitle
\begin{abstract}
In recent years, large language models (LLMs) have emerged as promising candidates for graph tasks. Many studies leverage natural language to describe graphs and apply LLMs for reasoning, yet most focus narrowly on performance benchmarks without fully comparing LLMs to graph learning models or exploring their broader potential. In this work, we present a comprehensive study of LLMs on graph learning tasks, evaluating both off-the-shelf and instruction-tuned models across a variety of scenarios. Beyond accuracy, we discuss data leakage concerns and computational overhead, and assess their performance under few-shot/zero-shot settings, domain transfer, structural understanding, and robustness. Our findings show that LLMs, particularly those with instruction tuning, greatly outperform traditional graph learning models in few-shot settings, exhibit strong domain transferability, and demonstrate excellent generalization and robustness. Our study highlights the broader capabilities of LLMs in graph learning and provides a foundation for future research \footnote{\url{https://github.com/myflashbarry/LLM-benchmarking}}. 
\end{abstract}

\section{Introduction}
\label{sec:Introduction}
The rapid progress of large language models (LLMs), such as GPTs \cite{achiam2023gpt4}, LLaMA \cite{touvron2023llama}, Claude \cite{claude}, and Deepseek \cite{liu2024deepseek}, has revolutionized many natural language processing tasks, showcasing their ability to generalize across domains and reason with minimal supervision. Recently, researchers have begun extending LLMs to non-text domains like graphs, aiming to leverage their strong reasoning capabilities for graph tasks.

Unlike text, graphs represent structured relational data, posing new challenges for LLMs in terms of representation and reasoning. To bridge this gap, various approaches have emerged: some utilize prompt engineering to describe graph structures in natural language \cite{cao2024graphinsight, zhang2024llm4dyg, kim2023kg, jiang2023structgpt, nlgraph, fatemi2023talklikeagraph}, while others integrate graph embeddings from graph neural networks (GNNs) or graph transformers (GTs) into LLMs \cite{chen2024llaga, chai2023graphllm, tang2024graphgpt, perozzi2024let}. To further mitigate the semantic gap between graphs and text, instruction tuning \cite{instructglm, tang2024graphgpt, zhang2023graph} is introduced, enabling LLMs to better understand graph features and structures. 

Meanwhile, graph learning models continue to evolve. Classic GNNs \cite{gcn, graphsage, gat, gin} rely on message passing and aggregation to capture local graph structures, but their performance often depends heavily on labeled data. To alleviate this reliance, graph self-supervised learning (SSL) methods \cite{you2020graphcl, dgi, hou2022graphmae} adopt a pre-training–fine-tuning paradigm, using unlabeled data to learn meaningful structural representations. In parallel, GTs \cite{graphormer, graphbert} have been proposed to overcome the locality constraints of GNNs by using self-attention to model long-range dependencies. More recently, foundational graph prompt models \cite{ofa, huang2024prodigy, sun2023allinone} have introduced the concept of graph prompts as a way to better align pre-trained models with downstream tasks, thereby enhancing generalization and adaptability.

However, existing studies on applying LLMs to graph learning tasks often adopt inconsistent experimental settings, including variations in datasets, preprocessing methods, and splitting strategies \cite{li2024glbench}. These inconsistencies hinder systematic comparison and obscure a clear understanding of how LLMs truly perform relative to graph learning models. To bridge this gap, we conduct a comprehensive evaluation of LLMs alongside 16 diverse graph learning models, encompassing GNNs, graph SSL, GTs, LM-augmented graph models, and foundational graph prompt methods. To ensure fairness and reproducibility, we standardize data processing pipelines and splitting protocols across graph datasets, covering both node classification and link prediction tasks. Our benchmark further includes a broad spectrum of LLMs, ranging from open-source models such as Llama3B and Llama8B to proprietary systems like Qwen-max, GPT-4o, Deepseek V3, and Gemini2.5 Pro.

Our benchmarking results (see details in Section~\ref{sec:benchmarking_results}) show that pure LLMs, especially larger LLMs, perform on par with or even surpass most baseline models in node classification and link prediction tasks. Instruction tuning further boosts LLM performance, enabling even smaller models to match or exceed the performance of top baseline models. Given the promising potential of instruction tuning, we further explore how LLMs with instruction tuning perform in other critical cases.

While instruction tuning significantly enhances LLM performance, it typically relies on abundant labeled data, which is often unavailable in real-world scenarios \cite{xia2024opengraph}. To assess its effectiveness under data scarcity, we first evaluate instruction-tuned LLMs in few-shot settings, examining whether they retain strong predictive capabilities with minimal supervision. We then explore their transferability across tasks and domains, a key property for practical deployment in low-resource environments. To further test their robustness, we introduce structural perturbations that commonly occur in real-world graphs such as missing node features, edge deletions, and reduced topological similarity. In addition, we examine potential data leakage, extend evaluation to more diverse datasets and graph classification task, and assess model performance on graph reasoning tasks such as shortest path and maximum flow. Unlike graph learning tasks (e.g., node/link/graph classification), which are probabilistic and data-driven, graph reasoning tasks are algorithmic with deterministic solutions that can be computed without learning. Evaluating both types provides a more comprehensive view of LLMs’ generalization, reasoning, and robustness in realistic graph settings.

\vspace{-2mm}
\paragraph{\textbf{Existing Benchmarks for LLMs in Graph Learning Tasks}}
There are some benchmarking works that explore the performance of LLMs on graph learning tasks. Studies like \cite{chen2024text} and \cite{yan2023comprehensive} focus on how LLMs can enhance graph models (e.g., GNNs) rather than benchmarking pure LLMs on graph learning tasks. GraphICL \cite{sun2025graphicl} aims to improve LLM performance in node classification and link prediction through various graph prompts, with an emphasis on prompt engineering, but it does not explore the impact of instruction tuning on LLMs in graph learning tasks. GLBench \cite{li2024glbench} also centers on how LLMs can better assist graph models, without focusing on purely LLM-based performance in graph learning tasks. Although both \cite{wu2025comprehensive} and \cite{zhu2024investigating} consider instruction-tuned LLMs, the former mainly examines their zero-shot capabilities and integration with graph models, without addressing the broader impact of instruction tuning or its role in link prediction. The latter does not thoroughly explore the performance of instruction-tuned LLMs under few-shot/zero-shot settings, domain transfer, structural understanding, and robustness—factors that are especially critical in data-scarce scenarios. \emph{To the best of our knowledge, our work provides one of the most comprehensive evaluations of pure LLMs on graph learning tasks incorporating instruction tuning. Moreover, we go beyond prior studies by systematically evaluating instruction tuning under practical data scarcity scenarios, providing a more thorough understanding of its impact on LLM performance in graph learning tasks.}

\begin{figure*}[htbp]
  \centering
  \includegraphics[width=1\linewidth]{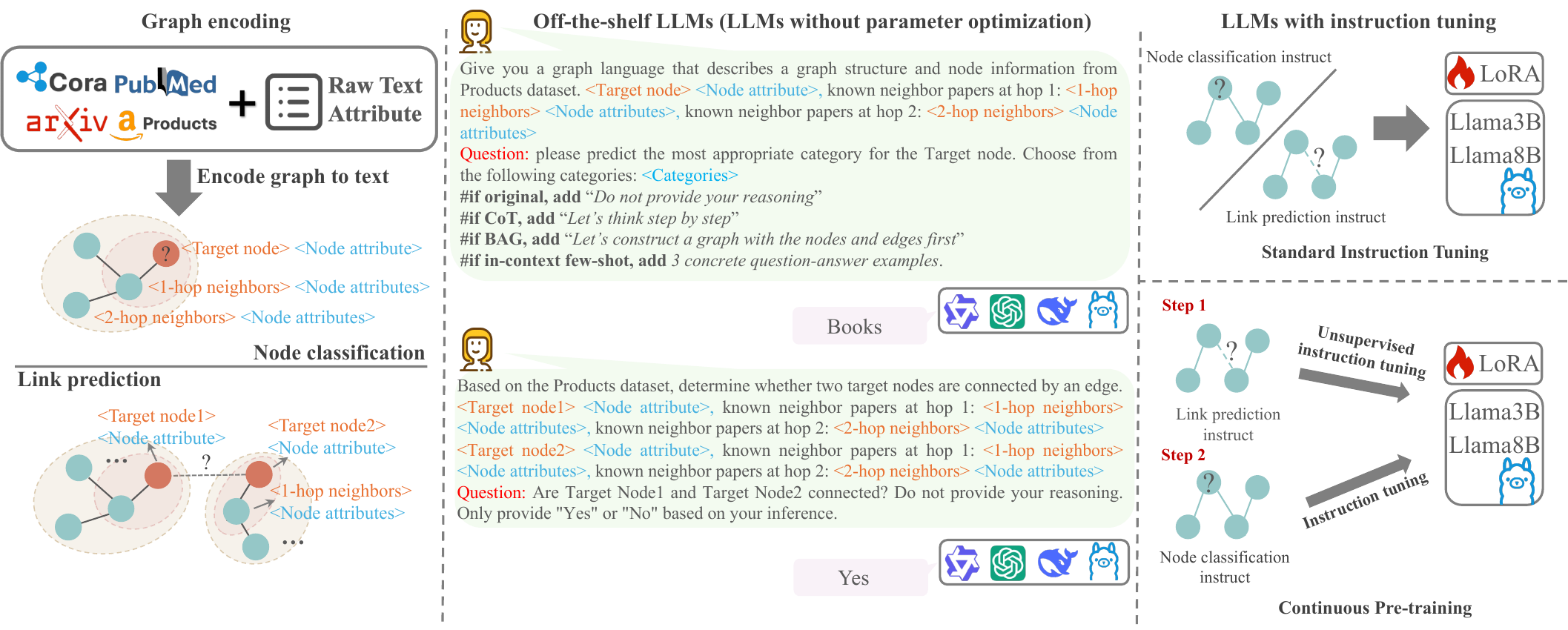} 
  \caption{The overall experimental pipeline for LLMs. Graph encoding outlines how prompts for LLMs are generated. Off-the-shelf LLMs show the question-answering process with LLMs. LLMs with instruction tuning describe the process of fine-tuning LLMs specifically for graph learning tasks.}
  \label{fig:The overall pipeline of our benchmark}
\vspace{-3mm}
\end{figure*}

\vspace{-2mm}
\section{Graph Learning with Pure LLMs}
\vspace{-2mm}
In this section, we introduce how we use LLMs for important real-world graph learning tasks, including node classification and link prediction.

\vspace{-2mm}
\subsection{Prompt Design}\label{sec:prompt_design}

As shown in the graph encoding part of Figure \ref{fig:The overall pipeline of our benchmark}, we combine the original graph datasets with their corresponding raw text attributes to encode the graph into a format that LLMs can understand, i.e., prompts. The prompt formats required for node classification and link prediction differ based on the specific task.
\vspace{-2mm}
\paragraph{\underline{Prompt formats for node classification}}
Following \cite{whenandwhy}, we adopt three basic prompt formats that use only the target node features, its 1-hop neighbors, or its 2-hop neighbors. In the original design, neighbor labels are included, which improves reasoning but may overly simplify the task by providing direct supervision. To better assess LLMs’ ability to learn from structure alone, we introduce two additional formats that exclude neighbor labels. In total, we evaluate five prompt formats, and detailed prompt structures are provided in Appendix~\ref{sec:Prompt Formats for Node Classification}:
\vspace{-1mm}
\begin{enumerate}[itemsep=0pt, parsep=0pt,leftmargin=*]
\item \textbf{ego}: Only the target node attributes.
\item \textbf{1-hop w/o label}: Attributes of the target node and its 1-hop neighbors, excluding labels.
\item \textbf{2-hop w/o label}: Attributes of the target node and its 2-hop neighbors, excluding labels.
\item \textbf{1-hop w label}: Same as above, but with 1-hop neighbor labels from the training set.
\item \textbf{2-hop w label}: Includes 2-hop neighbor labels from the training set.
\end{enumerate}

\vspace{-2mm}
\paragraph{\underline{Prompt formats for link prediction}}
We adopt two prompt formats to determine the existence of an edge between two target nodes: 1) \textbf{1-hop}: Both target nodes are described using their own node attributes and those of their 1-hop neighbors. 2) \textbf{2-hop}: extended to 2-hop neighbors. To avoid trivial cases, the two target nodes are never included in each other’s neighborhood. Full prompt examples are in Appendix~\ref{sec:Prompt Formats for Link Prediction}.

\vspace{-2mm}
\subsection{Paradigm of Using LLMs for Graph Learning Tasks}
\vspace{-2mm}
As shown in Figure~\ref{fig:The overall pipeline of our benchmark}, we explore two usage paradigms: (1) off-the-shelf LLMs, which are used without parameter updates, and (2) LLMs with instruction tuning. 

\vspace{-2mm}
\paragraph{\underline{Off-the-shelf LLMs}}
The LLMs we use are Llama-3.2-3B-Instruct (Llama3B), Llama-3.1-8B-Instruct (Llama8B), and the closed-source Qwen-plus \cite{bai2023qwen}. We directly evaluate them by feeding them carefully designed prompts encoding graph information and comparing their outputs to ground truth. Beyond basic prompts (Section~\ref{sec:prompt_design}), we experiment with Chain of Thought (CoT) \cite{CoT}, Build A Graph (BAG) \cite{nlgraph}, and in-context few-shot prompting on larger models like Qwen-max \cite{bai2023qwen}, GPT-4o \cite{achiam2023gpt4}, Deepseek V3 \cite{liu2024deepseek}, and Gemini2.5 Pro \cite{comanici2025gemini}. Results show that prompt strategies vary widely in effectiveness across datasets and model scales, and do not always lead to improvements. Full comparisons are provided in Appendix~\ref{sec:Comparison of Different LLMs on Node Classification}.

\vspace{-2mm}
% \paragraph{\textbf{LLMs with instruction tuning}}
\paragraph{\underline{LLMs with instruction tuning}}
We fine-tune Llama3B and Llama8B using LoRA \cite{hu2021lora}, with one training epoch per model, as longer training shows limited gains. For node classification, tuning is limited to the ego, 1-hop w/o label, and 2-hop w/o label formats. For link prediction, we examine both the benefits of tuning and the role of prompt diversity, an aspect underexplored in prior work. Two tuning modes are used: one aligns with the test formats (1-hop, 2-hop), and the other introduces nine diverse formats varying in question style and neighbor scope. Full prompt details are in Appendix~\ref{sec:Prompt Formats for Link Prediction}.

\vspace{-2mm}
\section{Comprehensive Benchmarking of LLMs for Graph Learning Tasks}
\label{sec:Fair Benchmark}

\vspace{-2mm}
Existing studies on LLMs for graphs \cite{tang2024graphgpt, zhao2023graphtext, li2024glbench} vary in datasets, preprocessing, and splitting, making comparisons difficult and obscuring LLM performance. Many works also benchmark against only limited baselines: \cite{yan2023comprehensive, instructglm} focus on classic GNNs and GTs, while \cite{tang2024graphgpt} considers only GNNs and graph SSL models, overlooking recent foundational graph prompt models (e.g., OFA \cite{ofa}). This narrow scope limits insight into LLM strengths and weaknesses in graph learning. Thus, we build a comprehensive benchmark covering a broader range of models for node classification and link prediction.

\vspace{-2mm}
\subsection{The Overall Setup} 
\label{sec:The Overall Setup}
\vspace{-1mm}
\subsubsection{Baselines}
% \ym{In this part, we xxx}
\vspace{-2mm}

For baseline models, we conduct a comprehensive comparison across 6 graph learning paradigms, covering a total of 16 graph models, including both traditional GNNs and more advanced architectures. This ensures a thorough evaluation of the capabilities of LLMs. The details about baseline models can be found in Appendix \ref{sec:Baseline models}.

\vspace{-2mm}
\subsubsection{Datasets}
\vspace{-2mm}
For both node classification and link prediction, we use the Cora \cite{cora}, PubMed \cite{pubmed}, OGBN-ArXiv \cite{ogb}, and OGBN-Products \cite{ogb} datasets. For baseline models, we use their original node features (Appendix \ref{sec:Impacts of Different Node Feature Embedding Methods} discusses the impact of different node feature embedding methods). For LLMs, we preprocess the raw data to transform the node attributes into textual representations. 
Detailed descriptions of the datasets and their splitting methods can be found in Appendix \ref{sec:Datasets}.

\vspace{-2mm}
\subsubsection{Evaluation Settings}
\vspace{-2mm}
For both node classification and link prediction, we consistently use accuracy as the evaluation metric, the same as \cite{chen2024llaga} and \cite{instructglm}. In the case of link prediction, where the ratio of positive to negative samples in the test set is 1:1, accuracy is a suitable measure. To select the best model, we perform hyperparameter tuning, as different hyperparameters may cause model performance to vary across datasets.  
Detailed experimental settings and the hyperparameter search ranges for each model are provided in Appendix \ref{sec:Detailed-Experimental-Settings}.

\vspace{-2mm}
\subsection{Results and Analysis}\label{sec:benchmarking_results}
\vspace{-2mm}
We present and analyze the performance of various models across node classification and link prediction tasks, providing insights into the strengths and weaknesses of LLMs.
\vspace{-2mm}
\paragraph{\underline{Node classification}}
% \vspace{-2mm}

Table \ref{tab:node_classification_results_LLM} summarizes the performance across different datasets. We make some observations:
\begin{itemize}[itemsep=0pt, parsep=0pt,leftmargin=*]
\vspace{-2mm}
% \vspace{-10pt}
\item Classic GNNs show consistent accuracy, while GIANT \cite{giant} and TAPE \cite{tape} outperform them by using language models for improved node representations. Multiple-hop prompts yield better results than simpler prompts, indicating that LLMs benefit from richer graph context.

\item Label information improves performance by strengthening the model decision-making process, similar to in-context learning.

\item For instruction-tuned LLMs, Llama3B/8B shows notable improvements, especially with multiple-hop prompts. Tuned Llama8B achieves the highest average score, surpassing LLaGA \cite{chen2024llaga} and setting a new benchmark.
\vspace{-2mm}
\end{itemize}

\begin{table}[htbp]
\centering
\vspace{-1mm}
\small
\scriptsize
\renewcommand{\arraystretch}{0.9}
\setlength{\tabcolsep}{0.7mm}

{%
\begin{tabular}{l c c c c c c}
\toprule
\rowcolor{gray!10}
\textbf{Model} & \textbf{Prompt} & \textbf{Cora} & \textbf{PubMed} & \textbf{ArXiv} & \textbf{Products} & \textbf{Avg}\\ 
\midrule
GCN & - & 88.19 & 88.00 & 69.90 & 82.30 & 82.10\\
GraphSAGE & - & \underline{\textbf{89.67}}  & 89.02 & 71.35 & 82.89 & 83.23\\
GAT & - & 88.38 & 87.90 & 68.69 & 82.10 & 81.77\\
GraphCL & - & 83.58 & 82.86 & 67.87 & 80.20 & 78.63\\
GraphMAE& - & 75.98 & 82.82 & 65.54 & 77.32 & 75.42\\
Graphormer& - & 81.20 & 88.05 & 71.99 & 81.75 & 80.75\\
Prodigy& - & 77.32 & 83.6 & 70.86 & 80.01 & 77.95 \\
OFA& - & 78.31 & 78.56 & 73.92 & 83.12 & 78.48\\
GIANT& - & 89.10 & 90.48 & 74.41 & \textbf{84.33}  & 84.58\\
TAPE& - & 88.12 & 91.92 & 73.99 & 83.11 & 84.29\\ 

LLaGA& - & 88.94 & \textbf{94.57}  & \underline{\textbf{76.25}}  & 83.98  & \textbf{85.94} \\
\midrule

\multirow{5}{*}{Llama3B} & ego & 24.72 & 63.20 & 23.10 & 40.80  & 37.96\\
& 1-hop w/o label & 39.48 & 64.50 & 29.50 & 53.00  & 46.62\\
& 2-hop w/o label & 49.63 & 69.90 & 29.50 & 56.10  & 51.28\\
& 1-hop w label & 77.49 & 70.90 & \textbf{66.00}  & 68.80  & 70.80\\
& 2-hop w label & \textbf{83.03}  & \textbf{72.00}  & 65.20 & \textbf{71.20}  & \textbf{72.86}\\
\midrule
\multirow{5}{*}{Llama8B} & ego & 43.39 & 77.80 & 59.35 & 50.12  & 57.92\\
& 1-hop w/o label & 58.35 & 73.07 & 61.85 & 59.85  & 63.28\\
& 2-hop w/o label & 62.84 & \textbf{83.29} & \textbf{68.33} & 59.60  & 68.52\\
& 1-hop w label & 82.97 & 81.55 & 68.08 & 71.07  & 75.92\\
& 2-hop w label & \textbf{84.79} & 82.54 & 64.09 & \textbf{77.06}  & \textbf{77.12}\\
\midrule

\multirow{3}{*}{tuned Llama3B} & ego & 67.08 & 89.28 & 66.58 & 65.59 & 72.13\\
& 1-hop w/o label & 82.04 & 90.02 & 71.32 & 73.07 & 79.11\\
& 2-hop w/o label & \textbf{85.04}  & \textbf{91.52} & \textbf{72.82} & \textbf{77.89} & \textbf{81.82}\\
\midrule
\multirow{3}{*}{tuned Llama8B} & ego & 77.31 & 92.36 & 65.59 & 73.74 & 77.25\\
& 1-hop w/o label & 84.54 & 93.90 & 69.33 & 80.33 & 82.03 \\
& 2-hop w/o label & \underline{\textbf{89.67}}  & \underline{\textbf{95.22}}  & \textbf{76.01}  & \underline{\textbf{84.51}}  & \underline{\textbf{86.35}} \\
\bottomrule
\vspace{-2mm}
\end{tabular}
\caption{Performance of different models on node classification. The \textbf{best} results in each category are highlighted. The \underline{underline} means the overall best result.}
\label{tab:node_classification_results_LLM}
}
\end{table}
% \vspace{-8pt}
\vspace{-2mm}
\paragraph{\underline{Link prediction}}
The results for link prediction are presented in Table \ref{tab:LLM performance on link prediction}. We make the following observations:
\begin{itemize}[itemsep=0pt, parsep=0pt, leftmargin=*]
\item For baseline models, OFA \cite{ofa} achieves the best, benefiting from LLM-derived edge features during pre-training.

\item Off-the-shelf Llama3B/8B lag behind most baselines, while the larger Qwen-plus matches or surpasses them, underscoring the importance of model scale for graph reasoning.

\item Instruction-tuned LLMs achieve the best link prediction results. Using 2-hop prompts consistently outperforms 1-hop prompts, and tuning with 9 diverse formats yields better performance than with only 2, highlighting the value of rich structural prompts for reasoning.

\end{itemize}

\vspace{-2mm}
\begin{remark}
   Although smaller off-the-shelf LLMs underperform most baseline models, their reasoning ability improves as the model size increases and graph structure information is incorporated. Instruction tuning further enhances LLM performance on graph learning tasks, with even smaller models achieving performance comparable to or better than the best baseline models, particularly when more diverse instructions are applied.
\end{remark}
% \vspace{-10pt}
\vspace{-2mm}

% \vspace{-10pt}
\begin{table}[htbp]
\centering
\small
\scriptsize
\renewcommand{\arraystretch}{1}
\setlength{\tabcolsep}{0.7mm}

% \vspace{-8pt}
{%
\begin{tabular}{l c c c c c c}
\toprule
\rowcolor{gray!10}
\textbf{Models} & \textbf{Prompts} & \textbf{Cora} & \textbf{PubMed} & \textbf{ArXiv} & \textbf{Products} & \textbf{Avg} \\ 
\midrule

GCN &- & 87.78 & 86.22 & 90.34 & 89.75 & 88.52 \\
GraphSAGE &- & 84.39 & 78.81 & 92.98 & 92.98 & 87.29 \\
GAT &- & 86.88 & 82.81 & 83.33 & 85.57 & 84.65 \\
GraphCL &- & 92.98 & 93.76 & 90.85 & 94.21 & 92.95 \\
GraphMAE &- & 82.01 & 75.71 & 85.24 & 88.32 & 82.82 \\
Prodigy &- & 90.9 & 91.67 & 89.22 & 92.99 & 91.2\\
OFA &- & 94.19 & 98.05 & 95.84  & 96.90  & 96.25 \\
LLaGA &- & 87.01 & 90.10 & 93.88 & 95.67 & 91.67 \\
\hline

Llama3B & 2-hop & 68.21 & 59.95 & 68.55 & 79.17 & 68.97 \\
Llama8B& 2-hop & 89.39 & 77.30 & 92.30 & 90.77 & 87.44 \\
Qwen-plus& 2-hop & 90.91 & 95.04 & 93.39 & 90.12 & 92.37 \\
\midrule
\multirow{2}{*}{t-Llama3B (2 formats)} & 1-hop & 83.12 & 93.95 & 92.20 & 90.07 & 89.84 \\
& 2-hop & \underline{95.76}  & 98.35 & 95.45 & 94.65 & 96.05 \\
\midrule
\multirow{2}{*}{t-Llama3B (9 formats)} & 1-hop & 87.18 & 94.40 & 93.30 & 95.45 & 92.58 \\
& 2-hop & \textbf{95.94}  & \textbf{99.20}  & 95.42 & \underline{97.84}  & \textbf{97.10}  \\
\midrule
\multirow{2}{*}{t-Llama8B (2 formats)} & 1-hop & 88.65 & 95.12 & 93.65 & 93.23 & 92.66 \\
& 2-hop & 95.39  & 98.77  & \textbf{96.11}  & 94.92 & 96.30  \\
\midrule
\multirow{2}{*}{t-Llama8B (9 formats)} & 1-hop & 88.47 & 96.01 & 95.21 & 96.33 & 94.01 \\
& 2-hop & 95.15 & \textbf{99.20}  & \underline{95.89}  & \textbf{97.98}  & \underline{97.06}  \\
\bottomrule
\end{tabular}
\caption{LLM performance on link prediction. The \textbf{best} and \underline{second-best} are highlighted. t-Llama3B means tuned Llama3B, t-Llama8B means tuned Llama8B.}
\label{tab:LLM performance on link prediction}
}

\end{table}

% \vspace{\baselineskip}
A theoretical justification for the effectiveness of instruction tuning is provided in Appendix~\ref{app:theoretical_justification}.

\begin{table*}[htbp]
\centering
\small
\scriptsize
\renewcommand{\arraystretch}{1}
\setlength{\tabcolsep}{0.65mm}
% \vspace{-8pt}

{%
\begin{tabular}{l c c c c c c | c c c c c | c c c c c}
\toprule
\rowcolor{gray!10}
 & & \multicolumn{5}{c}{\textbf{Full fine-tune}} & \multicolumn{5}{c}{\textbf{5-shot}} & \multicolumn{5}{c}{\textbf{10-shot}} \\
\cline{3-7} \cline{8-12} \cline{13-17}
\rowcolor{gray!10}
\textbf{Models} & \textbf{Prompts} & \textbf{Cora} & \textbf{PubMed} & \textbf{ArXiv} & \textbf{Products} & \textbf{Avg} & \textbf{Cora} & \textbf{PubMed} & \textbf{ArXiv} & \textbf{Products} & \textbf{Avg} & \textbf{Cora} & \textbf{PubMed} & \textbf{ArXiv} & \textbf{Products} & \textbf{Avg} \\ 
\midrule

GCN &- & 88.19 & 88.00 & 69.90 & 82.30 & 82.10 & 62.13 & 68.19 & 24.62 & 47.77 & 50.68 & 71.75 & 71.81 & 25.63 & 54.60 & 55.95 \\
GraphSAGE &- & \underline{\textbf{89.67}}  & \textbf{89.02}  & \textbf{71.35}   & \textbf{82.89}  & \textbf{83.23}  & 58.91 & 65.58 & 19.12 & 45.94 & 47.39 & 70.29 & 70.90 & 22.91 & 51.29 & 53.85 \\
GAT &- & 88.38 & 87.90 & 68.69 & 82.10 & 81.77 & 54.95 & 63.95 & 19.08 & 32.65 & 42.66 & 69.26 & 70.60 & 25.34 & 43.59 & 52.20 \\
GraphCL &- & 83.58 & 82.86 & 67.87 & 80.20 & 78.63 & 54.03 & 54.86 & 11.24 & 34.10 & 38.56 & 57.96 & 55.23 & 16.84 & 46.08 & 44.03 \\
GraphMAE &- & 75.98 & 82.82 & 65.54 & 77.32 & 75.42 & 24.44 & \textbf{70.47}  & 24.26 & 50.61 & 42.45 & 30.59 & \textbf{73.63} & 28.64 & 57.55 & 47.60 \\
All in one &- &- &- &- &- &- & 50.98 & 60.49 & 16.34 & 41.18 & 42.25 & 51.66 & 61.93 & 20.42 & 47.73 & 45.44 \\
GPF-plus &- &- &- &- &- &- & \textbf{67.00}  & 66.91 & 60.07 & \textbf{64.50}  & 64.62 & \textbf{73.22} & 64.39 & 65.35 & \textbf{68.02} & 67.75 \\
GraphPrompt &- &- &- &- &- &- & 65.12 & 68.11 & \underline{\textbf{81.88}}  & 58.44 & \textbf{68.39}  & 69.81 & 70.38 & \underline{\textbf{87.05}}  & 61.02 & \textbf{72.07}  \\
\midrule

Llama3B & 2-hop w/o label & 85.04  & 91.52  & 72.82  & 77.89  & 81.82  & 76.81  & 71.32  & 55.24  & 67.32  & 67.67  & 77.81  & 85.53  & 63.33  & 68.11  & 73.70  \\
Llama8B & 2-hop w/o label & \textbf{89.67}  & \underline{\textbf{95.22}}  & \underline{\textbf{76.01}}  & \underline{\textbf{84.51}}  & \underline{\textbf{86.35}}  & \underline{\textbf{77.10}}  & \underline{\textbf{79.43}}  & \textbf{69.78}  & \underline{\textbf{73.12}}  & \underline{\textbf{74.86}}  & \underline{\textbf{80.55}}  & \underline{\textbf{88.89}}  & \underline{\textbf{71.12}}  & \underline{\textbf{74.86}}  & \underline{\textbf{78.86}}  \\
\bottomrule
\end{tabular}
\caption{Performance of models under few-shot learning. The \textbf{best} results in each category are highlighted. The \underline{underline} means the overall best result.}
\label{tab:node_classification_results_few_shot_LLM}
}
\vspace{-3mm}
\end{table*}
% \section{Further Investigation }

\vspace{-2mm}
\section{Further Investigation on LLMs with Instruction Tuning}
\label{sec:Research Questions}
\vspace{-2mm}
Instruction tuning enables even small LLMs to perform well, but data scarcity remains a major challenge in real-world scenarios \cite{xia2024opengraph}. Traditional graph models like GNNs and graph transformers often suffer under limited labeled data due to their reliance on structural and label information \cite{yu2024survey}. Recent models such as All in One \cite{sun2023allinone} and GPF-plus \cite{gpf-plus} aim to improve performance in low-label settings, yet the behavior of instruction-tuned LLMs under such constraints is still underexplored. Therefore, in this section, we discuss methods to alleviate data scarcity and further explore the performance of LLMs with instruction tuning in such scenarios.

Label scarcity is a common data limitation. Improving few-shot performance remains a key goal for both graph models \cite{yu2024survey, zhao2024pre} and LLMs. For LLMs, few-shot instruction tuning sheds light on their robustness to label scarcity and their ability to generalize from limited supervision—crucial for real-world applicability. This motivates the following research question:

% \vspace{0.5\baselineskip}
\begin{mdframed}[backgroundcolor=gray!8]
\textbf{\textit{RQ1: How well do LLMs perform in few-shot instruction tuning scenarios?}}
\end{mdframed}
% \vspace{0.5\baselineskip}

When labeled data is limited, leveraging unlabeled data is a natural way to improve model performance. This idea underlies continual learning, where models incrementally adapt to new data with minimal supervision \cite{wang2024comprehensive, van2019three}. For LLMs, continual domain-adaptive pre-training~\cite{ke2023continual, yildiz2024investigating} has proven effective for enhancing downstream performance. Inspired by this, we propose continuous pre-training for graph tasks, where LLMs are first unsupervisedly trained on graph-structured data, then fine-tuned with task-specific data. As unlabeled graph data is far more abundant, this approach holds promise for improving LLM adaptability. This motivates the following research question:

% \vspace{0.5\baselineskip}
\begin{mdframed}[backgroundcolor=gray!8]
\textbf{\textit{RQ2: How does continuous pre-training impact the performance of LLMs?}}
\end{mdframed}
% \vspace{0.5\baselineskip}

Models with strong transferability can mitigate performance drops under label scarcity by transferring knowledge from other datasets. LLMs have shown impressive transferability in natural language tasks \cite{du2024unlocking, ran2024alopex}, but their transferability in graph tasks has been less explored. If instruction-tuned LLMs can generalize well across different graph domains, a one-time tuning process could support multiple downstream tasks, greatly reducing resource costs. This raises the following research question:

% \vspace{0.5\baselineskip}
\begin{mdframed}[backgroundcolor=gray!8]
\textbf{\textit{RQ3: How well do LLMs transfer knowledge across domains in node classification and link prediction?}}
\end{mdframed}
% \vspace{0.5\baselineskip}

\paragraph{\textbf{Further Probing}}
Given the central role of structural information in graph learning tasks, along with the practical challenges posed by real-world perturbations (e.g., missing edges) and the high computational cost of LLMs, we further investigate three key aspects: structural understanding, robustness to structural noise, and computational efficiency. Detailed analyses are provided in Appendix~\ref{sec:Case 5}, \ref{sec:Robustness of LLMs}, and~\ref{sec:Computational Overhead Analysis}.

\vspace{-2mm}
\section{Experiment and  Analysis}
\label{sec:Empirical Studies}
\vspace{-2mm}
We conduct empirical studies on different research questions proposed in Section \ref{sec:Research Questions}. 
In the following subsections, we first introduce the experimental settings for each RQ, followed by experimental results analysis and key remarks.

\vspace{-2mm}
\subsection{Few-Shot Instruction Tuning (RQ1)}
\label{sec:Case 2}
\subsubsection{Experiment Settings}
We focus on few-shot instruction tuning for node classification. We use 2-hop w/o label as prompt formats and randomly select 5 or 10 target nodes per class for instruction tuning.
For baselines, we include not only GNNs and graph SSL models but also foundational prompt-based methods including All in One~\cite{sun2023allinone}, GPF-plus~\cite{gpf-plus}, and GraphPrompt~\cite{liu2023graphprompt}, which leverage pre-trained knowledge and graph prompts to perform well in few-shot settings.

\vspace{-2mm}
\subsubsection{Results}

Table \ref{tab:node_classification_results_few_shot_LLM} summarizes the results. All models experience a decline in accuracy under few-shot learning compared to full fine-tuning, with GNNs and Graph SSL models showing the largest drops, particularly in larger datasets like ArXiv and Products. In contrast, LLMs exhibit more consistent performance, indicating greater robustness in data-scarce scenarios. Notably, Llama8B achieves the highest classification accuracy in both 5-shot and 10-shot scenarios, showing LLMs’ ability to learn quickly from limited data.

\vspace{-2mm}
\begin{remark}
LLMs outperform all other models in few-shot scenarios. Only a few foundational graph prompt models achieve comparable results on certain datasets, underscoring LLMs’ clear advantage in data-scarce situations.
\end{remark}
\vspace{-2mm}

\vspace{-2mm}
\subsection{Impact of Continuous Pre-training (RQ2)}
\label{sec:Case 3}

\vspace{-2mm}
As shown in Figure~\ref{fig:The overall pipeline of our benchmark}, continuous pre-training involves two stages. First, the model undergoes task-agnostic unsupervised learning on the target dataset to acquire general graph representations. It is then instruction-tuned on a task aligned with the inference objective.
\vspace{-2mm}
\subsubsection{Experiment Settings}
\vspace{-2mm}

We evaluate both zero-shot and few-shot node classification. In the zero-shot setting, the model is first continuously pre-trained via unsupervised link prediction, then directly evaluated on node classification. Baselines include LLaGA and ZeroG~\cite{li2024zerog}, a prompt-based model tailored for zero-shot tasks. In the few-shot setting, we apply instruction tuning either with or without the preceding link prediction step for comparison.

\begin{table}[htbp]
\centering
\small
\scriptsize
\renewcommand{\arraystretch}{1}
\setlength{\tabcolsep}{0.65mm}

{%
\begin{tabular}{l c c c c c c}
\toprule
\rowcolor{gray!10}
\textbf{Models} & \textbf{Prompts} & \textbf{Cora} & \textbf{PubMed} & \textbf{ArXiv} & \textbf{Products} & \textbf{Avg} \\ 
\midrule
ZeroG &- & 68.61 & 78.77 & 70.50   & 55.23 & 68.28 \\
LLaGA &- & 22.03 & 55.92 & 21.15 & 38.90 & 34.50 \\
\midrule

Llama3B& 2-hop & 49.63 & 69.90 & 29.50 & 56.10 & 51.28 \\

Llama3B w CPT& 2-hop & 55.36 & 75.56 & 33.54 & 57.01 & 55.37 \\

Llama3B w 5s & 2-hop & 76.81 & 71.32 & 55.24 & 67.32 & 67.67  \\

Llama3B w CPT \& 5s& 2-hop & \textbf{79.58}   & \underline{88.53}   & 54.11 & 68.08 & 72.58  \\
\midrule

Llama8B& 2-hop & 62.84 & 83.29 & 68.33 & 59.60 & 68.52 \\

Llama8B w CPT& 2-hop & 70.82  & 86.96   & \textbf{71.34}   & 63.20 & 73.08  \\

Llama8B w 5s& 2-hop & 77.1 0  & 79.43 & 69.78  & \underline{73.12}   & 74.86   \\

Llama8B w CPT \& 5s& 2-hop & \underline{78.12}   & \textbf{89.03}   & \underline{71.01}   & \textbf{74.69}   & \textbf{78.21}   \\
\bottomrule
\end{tabular}
\caption{Performance of continuous pre-training for LLM. "w CPT" means zero-shot inference after continuous pre-training. "w 5s" means direct 5-shot instruction tuning without continuous pre-training. "w CPT \& 5s" means 5-shot instruction tuning after continuous pre-training. The \textbf{best} and \underline{second-best} are highlighted.}
\label{tab:continuous pre-training for LLM}
}
\vspace{-2mm}
\end{table}

\vspace{-2mm}
\subsubsection{Results}
\vspace{-1mm}

Table~\ref{tab:continuous pre-training for LLM} shows that continuous pre-training improves LLM performance over zero-shot (e.g., ZeroG) and few-shot learning, highlighting its effectiveness in enhancing graph understanding. On smaller datasets like Cora and PubMed, Llama3B with continuous pre-training matches Llama8B. However, on larger datasets such as Arxiv and Products, LLaMA-8B still leads, suggesting that model scaling remains crucial for complex graphs.

\vspace{-1mm}
\begin{remark}
Continuous pre-training can significantly improve LLM performance in zero-shot and few-shot learning. However, for larger and more complex datasets, increasing the size of the LLM proves to be a more effective approach.
\end{remark}
\vspace{-2mm}
\vspace{-2mm}
\subsection{Domain Transferability of LLMs (RQ3)}
\label{sec:Case 4}
\vspace{-1mm}

Domain transferability can be divided into in-domain (across datasets within the same domain) and cross-domain (across different domains) based on difficulty. We evaluate LLMs with instruction tuning in both settings.

\vspace{-2mm}
\subsubsection{Experiment Settings}
\vspace{-2mm}

In the in-domain setting, we train on the Arxiv citation graph and evaluate on Cora, another citation dataset. For cross-domain transfer, we train on Arxiv and test on Products, an e-commerce graph. GNNs rely on task-specific heads, limiting their zero-shot capability when label sets differ, so we focus on LLaGA for node classification. For link prediction, we apply a simple linear mapping to align feature dimensions across datasets. Baselines include GNNs, graph SSL models, and LLaGA.

\begin{figure}[htbp]
  \centering
  \vspace{-2mm}
  \includegraphics[width=1\linewidth]{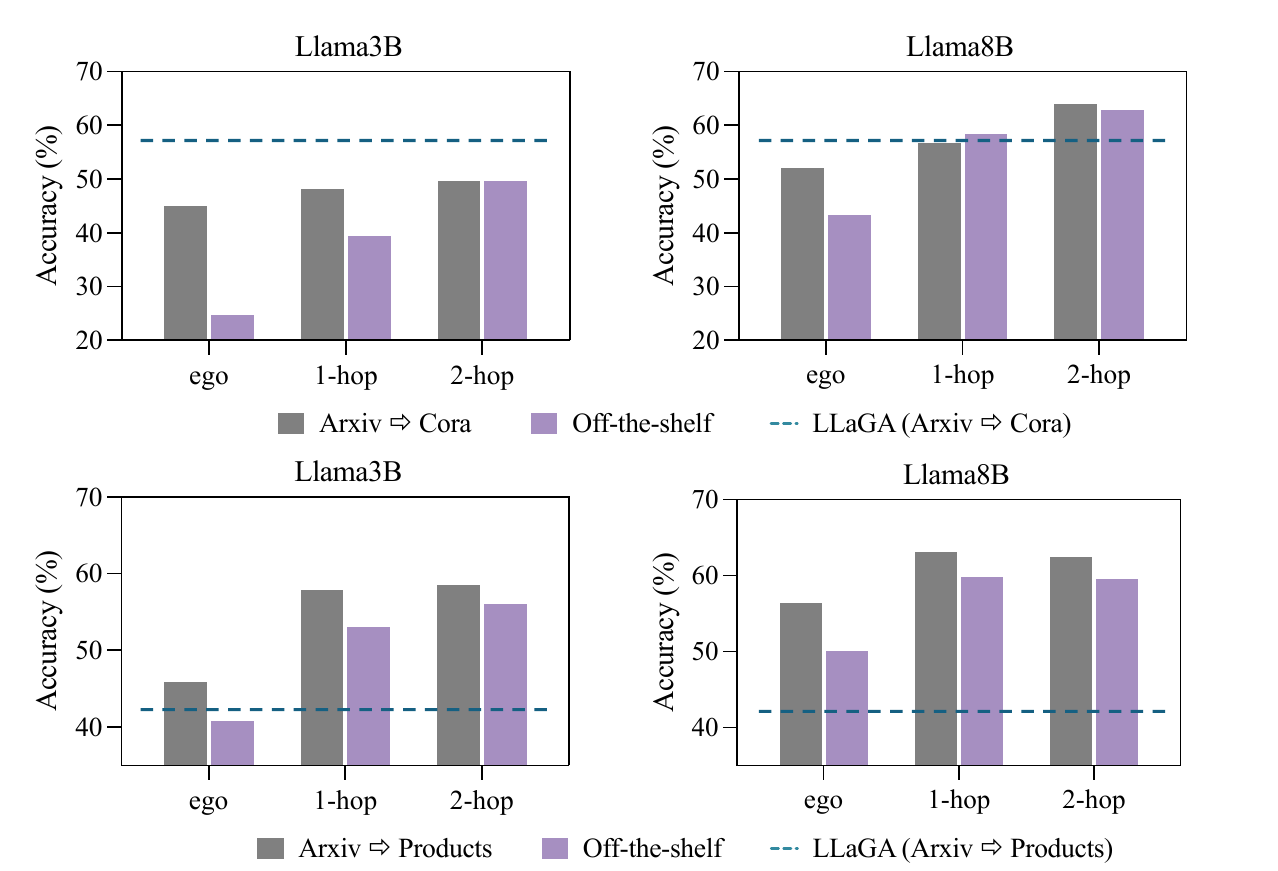}
  %\vspace{-20pt}
  \caption{LLM domain transferability in node classification}
  \label{fig:Domain Transferability}
  \vspace{-2mm}
\end{figure}

% \vspace{-10pt}
\vspace{-2mm}
\subsubsection{Results}

\paragraph{\underline{Node classification}}

Figure \ref{fig:Domain Transferability} presents the accuracy of different models in both in-domain and cross-domain scenarios. Instruction-tuned LLMs on Arxiv outperform off-the-shelf scenario, but the improvement is modest when additional structural information is incorporated. This is likely due to the fact that node classification relies heavily on category information, and adding more structural data does not significantly enhance performance. While LLMs learn graph information from Arxiv, adapting to unseen categories remains challenging, limiting performance gains. Besides, LLMs perform comparably to LLaGA on Cora dataset, but on the more complex Products dataset, LLMs show a clear advantage.
This suggests that the simple graph projector of LLaGA struggles to capture diverse graph patterns, while LLMs can adapt better to varying structures and are capable of learning diverse feature information with their sophisticated instruction tuning mechanisms.

\begin{table}[htbp]
\vspace{-2mm}
\centering
\small
\scriptsize
\renewcommand{\arraystretch}{0.9}
% \setlength{\tabcolsep}{1mm}

% \vspace{-10pt}
{%
\begin{tabular}{l c c| c }
\toprule
\rowcolor{gray!10}
 & & \multicolumn{2}{c}{\textbf{Train $\longrightarrow$ Test}} \\
\cline{3-4}
\rowcolor{gray!10}
\textbf{Models} & \textbf{Prompts} & \textbf{Arxiv $\longrightarrow$ Cora} & \textbf{Arxiv $\longrightarrow$ Products} \\
\midrule
GCN &- & 55.54 & 67.07 \\
GraphSAGE &- & 50.00 & 51.11 \\
GAT &- & 85.41 & 71.18 \\
GraphCL &- & 78.30 & 82.62 \\
GraphMAE &- & 71.90 & 73.94 \\
LLaGA &- & 86.98 & 92.82 \\
\midrule
\multirow{2}{*}{Llama3B} & 1-hop & 87.55 & 91.16 \\
& 2-hop & \textbf{95.11}  & \underline{94.15}  \\
\midrule
\multirow{2}{*}{Llama8B} & 1-hop & 88.98 & 91.97 \\
& 2-hop & \underline{94.78}  & \textbf{95.43}  \\
\bottomrule
\end{tabular}
\caption{LLM domain transferability in link prediction. The \textbf{best} and \underline{second-best} are highlighted.}
\label{tab:LLM domain transferability}
}
\vspace{-2mm}
\end{table}

% \vspace{-10pt}
\vspace{-2mm}
\paragraph{\underline{Link prediction}}

From Table \ref{tab:LLM domain transferability}, we observe that LLMs significantly outperform traditional graph models. Only LLaGA achieve comparable performance, likely because it also leverages LLMs for predictions. 
In the in-domain transfer scenario, LLMs achieve performance on Cora comparable to models directly instruction-tuned on Cora, indicating they can effectively transfer knowledge from larger datasets to downstream tasks. In the cross-domain scenario, although LLM performance on Products is slightly lower than direct tuning, it still remains strong, possibly due to shared topological patterns across domains.

\vspace{-1mm}
\begin{remark}
LLMs with instruction tuning exhibit strong domain transferability, particularly in link prediction tasks, where they effectively generalize across different datasets. This may be because link prediction tasks across domains share more similarities, as they can be viewed as binary classification problems. In contrast, node classification is more challenging, as adapting learned knowledge to unseen categories is difficult. 
\end{remark}
\vspace{-2mm}

\vspace{-1mm}
\section{Further Probing}
\label{sec:further probing}
\vspace{-2mm}
To deepen our analysis, we extend evaluation to more diverse datasets and graph classification task, examine potential data leakage during pretraining, and assess model performance on graph reasoning tasks such as shortest path.
\vspace{-1mm}
\subsection{Broader Dataset and Task Evaluation}
\label{sec:graph classification}
\vspace{-2mm}
To strengthen the reliability of our conclusions, we conduct additional node classification experiments on four datasets and further evaluate graph classification tasks. As shown in Table~\ref{tab:More Datasets}, the results remain consistent with Section~\ref{sec:Fair Benchmark}: incorporating graph structure significantly improves LLM performance, and instruction tuning further enhances their effectiveness on graph learning tasks.

\begin{table}[htbp]
\vspace{-2mm}
\centering
\small
\scriptsize
\renewcommand{\arraystretch}{1}
\setlength{\tabcolsep}{0.3mm}

{%
\begin{tabular}{l c c c c c | c c}
\toprule
\rowcolor{gray!10}
\textbf{Model} & \textbf{Prompt} & \textbf{Computer} &\textbf{WikiCS} &\textbf{Reddit} &\textbf{Instagram} &\textbf{IMDB-B} & \textbf{HIV}\\ 
Task type & node & node &node &node &node &graph & graph\\ 
\midrule
GCN & - &  88.28 & 82.95 & 65.31 & 64.32 & 74.01 & 75.12\\

GraphCL & - &  76.52 & 84.84  & 62.19 & 63.10  & 71.16 & 77.36\\
Graphormer& - &  77.61 & \textbf{86.17}  & 66.30 & 62.32  &  73.28 & \underline{79.62}\\
OFA& - &  87.70 & 78.52 & 63.91 & 61.70  &  \underline{76.38} & 77.99\\
TAPE& - &  \underline{89.70}  & 83.60 & 63.97 & 65.11  & - & -\\ 

LLaGA& -  & \textbf{90.11}  & 80.01 & \textbf{67.10}  & \underline{66.60}   & - & - \\

\midrule
\multirow{3}{*}{Llama8B} & ego &  55.72 & 39.72 & 42.10 & 39.02 & - & -\\
& 2-hop &  67.00 & 70.31 & 51.92 & 46.60 & - & -\\
& - & - & - & - & - & 68.51 & 67.12 \\

\midrule
\multirow{3}{*}{t-Llama8B} & ego &  69.10 & 73.82 & 58.19 & 57.03 & - & -\\
& 2-hop &  86.40 & \underline{85.31}  & \underline{66.62}  & \textbf{68.31}  & - & - \\
& - & - & - & - & - & \textbf{83.01} & \textbf{85.13} \\
\bottomrule
\end{tabular}
\caption{Node classification of different models on more datasets. The \textbf{best} and \underline{second-best} are highlighted.}
\label{tab:More Datasets}
}
\vspace{-2mm}
\end{table}

\vspace{-2mm}
\subsection{Data Leakage Analysis}
\label{sec:data leakage}
\vspace{-1mm}
A key concern is the benchmark datasets may have been seen during LLM pre-training, posing a risk of data leakage. Following \cite{whenandwhy}, we compare ogbn-arxiv with arxiv-2023, a newly collected dataset of CS papers from year 2023. The two share similar citation structures, with aligned in-/out-degree distributions. We evaluate LLaMA-2-13B (trained on data up to September 2022) \cite{touvron2023llama} on node classification over both datasets, using the same setup as in Section~\ref{sec:Fair Benchmark}.

Table~\ref{tab:data leakage} presents the results. If data leakage were a key factor, LLMs would show a larger accuracy drop on the leakage-free arxiv-2023 than baseline models. However, the accuracy gap is comparable across models, and in some cases, LLMs even perform better on arxiv-2023. Graph structure and instruction tuning remain the dominant contributors to performance gains. These results suggest that any potential leakage has minimal impact, and that LLMs generalize well across datasets with different temporal distributions.

\begin{table}[htbp]
\vspace{-2mm}
\centering
\small
\renewcommand{\arraystretch}{0.75}
\scriptsize

{%
\begin{tabular}{l c c c}
\toprule
\rowcolor{gray!10}
\textbf{Model} & \textbf{Prompt} & \textbf{ogbn-arxiv} & \textbf{arxiv-2023}\\ 
\midrule
GCN & - & 69.90 & 65.33 \\

GraphCL & - &  67.87 & 66.82 \\
Graphormer& - & \textbf{71.99}  & \textbf{69.08}  \\
\midrule

\multirow{3}{*}{Llama13B} & ego &  55.24 & 57.70 \\
& 1-hop w/o label &  59.03 & 59.42  \\
& 2-hop w/o label &  65.90 & 63.02  \\

\midrule
\multirow{3}{*}{tuned Llama13B} & ego & 66.17 & 65.20\\
& 1-hop w/o label & 75.45 & \underline{\textbf{76.01}} \\
& 2-hop w/o label & \underline{\textbf{76.51}} & 75.82 \\

\bottomrule
\end{tabular}
\caption{Performance of different models on node classification tasks. The datasets are ogbn-arxiv and arxiv-2023. The \textbf{best} results in each category are highlighted. The \underline{underline} means the overall best result.}
\label{tab:data leakage}
}
\vspace{-2mm}
\end{table}
\vspace{-2mm}
\subsection{Graph Reasoning Task Evaluation}
\label{sec:Graph Reasoning Task Evaluation}
\vspace{-1mm}
While our main focus is benchmarking LLMs on data-driven graph learning tasks such as node classification and link prediction—which involve learning from data, handling uncertainty, and generalizing to unseen structures—LLMs can also be applied to graph reasoning tasks like shortest path and maximum flow. Unlike learning tasks, these problems have deterministic solutions via classical algorithms and require no training. Prior work \cite{nlgraph, zhang2024can, dai2024large, luo2024reasoninggraphsfaithfulinterpretable} has examined LLMs in this setting; here, we further study the effect of instruction tuning. We generate random graphs with 10–20 nodes using NetworkX \cite{hagberg2008exploring} and use the NLGraph \cite{nlgraph} prompt format to evaluate LLMs on 4 tasks: connectivity, circle, shortest path, and maximum flow.

\begin{table}[htbp]
\vspace{-2mm}
\centering
\small
\renewcommand{\arraystretch}{1}
\scriptsize
\setlength{\tabcolsep}{1.1mm}

{%
\begin{tabular}{l c c c c}
\toprule
\rowcolor{gray!10}
\textbf{Model} & \textbf{Connectivity} & \textbf{Cycle} & \textbf{Shortest path} & \textbf{Maximum flow}\\ 
\midrule
Random &50&	50	&6.22	&2.54 \\
% \midrule
Llama3B &56.48&	42.37&	2.52&	2.49\\
Llama8B& 63.97&	53.89&	11.24	&4.01\\
Qwen-max &72.85	&67.27&	30.07	&11.90\\
GPT-4o& 78.25	&66.75	&\underline{41.69}	&\underline{13.72}\\
Gemini2.5 Pro &\underline{83.13}	&73.15	&39.22	&\textbf{16.38}\\
tuned Llama3B& 74.73&	\underline{79.77}&	25.58	&11.85\\
tuned Llama8B& \textbf{86.40}&	\textbf{88.42}&	\textbf{56.98}&	13.47\\

\bottomrule
\end{tabular}
\caption{Performance of different models on graph reasoning tasks. The \textbf{best} results in each category are highlighted. The \underline{underline} means the overall best result.}
\label{tab:graph reasoning tasks}
}
\vspace{-2mm}
\end{table}

Table~\ref{tab:graph reasoning tasks} shows that closed-source LLMs perform reasonably well on graph reasoning tasks, and instruction tuning further improves their understanding—consistent with observations from graph learning tasks. However, for reasoning problems, classical algorithms (e.g., Dijkstra \cite{dijkstra2022note} for shortest path) remain more effective and reliable. In contrast, applying LLMs to graph learning tasks is more practical and meaningful in real-world settings.

\vspace{-2mm}
\section{Conclusion}
\label{sec:conclusion}
\vspace{-2mm}

This paper demonstrates that LLMs, especially with instruction tuning, achieve strong performance and surpass most graph models on graph learning tasks through a fair and comprehensive benchmarking approach. Our findings emphasize the potential of LLMs in few-shot learning, transferability, and understanding graph structures in data-scarce scenarios. The introduction of continuous pre-training further boosts LLM performance in such environments. These insights provide valuable guidance for the future application of LLMs in graph tasks, paving the way for more efficient and adaptable graph learning models in real-world settings.

\section{Limitations}

Our work, while comprehensive, has certain limitations that open avenues for future research. Firstly, our investigation primarily centers on node-level and link-level predictive tasks using text-attributed graphs, where LLMs can naturally leverage their semantic processing capabilities. The performance of pure LLM approaches on graph-level tasks (e.g., graph classification and regression) or on graphs with non-semantic, numerical, or sparse features is less explored. The generalization of our findings to these contexts remains an important open question.

Secondly, scalability presents a practical challenge. The LLM-based methods, particularly those involving instruction tuning, incur significantly higher computational costs than traditional GNNs. Furthermore, the approach of serializing graph neighborhoods into textual prompts is inherently constrained by the finite context window of LLMs. This may pose difficulties when applied to graphs with extremely large or dense neighborhoods, highlighting a need for future work on more efficient graph-to-text representations or sampling strategies.

\section{Ethical Considerations}

The datasets utilized in our study, such as Cora and PubMed, are publicly available benchmarks for academic research, minimizing direct ethical risks. However, we acknowledge broader ethical implications inherent in applying LLMs to graph data. Potential risks include the inference of sensitive user information from graph structures and the amplification of societal biases present in the textual data. Additionally, the significant computational resources required for training and fine-tuning these models contribute to a considerable environmental footprint. Future research should prioritize the development of privacy-preserving techniques and more computationally efficient models to mitigate these concerns.

\bibliography{custom}

\appendix

\section{Comparison between our benchmark and existing works}
In Table \ref{tab:comparison_existing_works}, we summarize the key differences between our benchmarking study and other papers. \textbf{Comprehensive Baselines} refers to whether the baseline models cover a wide range of model types. In our paper, we include GNNs, Graph SSL models, Graph Transformers, Foundational Graph Prompt Models, and LLMs with Graph Projectors. \textbf{Comprehensive Settings} examines the performance of models across various scenarios, such as vanilla fine-tuning, few-shot learning, and zero-shot learning. \textbf{Diverse LLMs} highlights the use of multiple LLMs for comparison, such as Llama, GPT, and Qwen. \textbf{LLM Tuning} indicates whether the paper fine-tunes the LLMs or simply uses the original models as they are. Lastly, \textbf{Transferability Study} explores whether the paper investigates the cross-task or cross-domain transfer capabilities of the models.

\begin{table*}[htbp]
\centering

\scalebox{0.51}{ % Adjust the scale as needed
\begin{tabular}{@{}lccccccc@{}}
\toprule
\rowcolor{gray!10}
\textbf{Model} & \textbf{Node Classification} & \textbf{Link Prediction} & \textbf{Comprehensive Baselines} & \textbf{Comprehensive Settings} & \textbf{Diverse LLMs} & \textbf{LLM Tuning} & \textbf{Transferability Study} \\ \midrule
InstructGLM \cite{instructglm}     & \ding{51} & \ding{55} & \ding{55} & \ding{55} & \ding{55} & \ding{51} & \ding{55} \\
LLaGA \cite{chen2024llaga}          & \ding{51} & \ding{51} & \ding{55} & \ding{55} & \ding{51} & \ding{55} & \ding{51} \\
InstructGraph \cite{wang2024instructgraph}   & \ding{51} & \ding{51} & \ding{55} & \ding{55} & \ding{51} & \ding{51} & \ding{55} \\
NLGraph \cite{nlgraph}        & \ding{55} & \ding{55} & \ding{55} & \ding{51} & \ding{55} & \ding{55} & \ding{55} \\
Talk Like a Graph \cite{fatemi2023talklikeagraph} & \ding{51} & \ding{55} & \ding{55} & \ding{51} & \ding{55} & \ding{55} & \ding{55} \\
All in One \cite{sun2023allinone}      & \ding{51} & \ding{51} & \ding{51} & \ding{51} & \ding{55} & \ding{55} & \ding{51} \\
OFA \cite{ofa}            & \ding{51} & \ding{51} & \ding{51} & \ding{51} & \ding{55} & \ding{55} & \ding{51} \\
GraphGPT \cite{tang2024graphgpt}       & \ding{51} & \ding{51} & \ding{55} & \ding{55} & \ding{51} & \ding{51} & \ding{51} \\ \midrule
\rowcolor{gray!10}
\textbf{Ours}   & \ding{51} & \ding{51} & \ding{51} & \ding{51} & \ding{51} & \ding{51} & \ding{51} \\ 
\bottomrule
\end{tabular}

}
\caption{Comparison between our benchmark and existing works}
\label{tab:comparison_existing_works}
\end{table*}

\section{Related Works}
\label{sec:related works}
In this section, we review the existing literature on the application of LLMs and related techniques in graph tasks. We highlight two primary categories: the use of LLMs for graph reasoning and their integration with traditional graph models to enhance performance.

\subsection{Large Language Models for Graph Reasoning}
Recent studies suggest that LLMs have the potential to solve graph reasoning tasks by understanding graph structures~\cite{fatemi2023talklikeagraph,tang2024grapharena}. NLGraph~\cite{nlgraph} indicates that LLMs can track paths within graphs, enabling them to solve tasks such as node connectivity and shortest path detection. Moreover, \citep{dai2024large} suggests that LLMs understand graph pattern concepts, which are fundamental to graph structure mining and learning. 

Additionally, fine-tuning further enhances the LLMs' reasoning ability in graph tasks~\cite{dai2024revisiting}. \cite{zhang2024can} suggest that LLMs can transfer their understanding of substructures through fine-tuning on graphs with different node features. Besides, GraphWiz~\cite{chen2024graphwiz} indicates that LLMs learn path reasoning across various tasks and datasets. Along the same line, FiDeLiS~\cite{sui-etal-2025-fidelis} leverages stepwise reasoning grounded in knowledge graphs to improve factual consistency in KG-QA, while Think-on-Graph 2.0~\cite{ma2025thinkongraph20deepfaithful} introduces iterative graph–context retrieval that strengthens LLM reasoning over complex structures. Furthermore, GraphGPT-O~\cite{fang2025graphgptosynergisticmultimodalcomprehension} extends graph reasoning to multimodal attributed graphs, enabling joint image–text generation with structural awareness. These advances highlight that LLMs can be effectively adapted to deepen their comprehension of graph structures.

\subsection{Language Model Aided Graph Models}
With the advancement of language models, their presence in graph-related tasks has become increasingly prominent. Their natural strengths in language processing and intrinsic reasoning make them particularly valuable, especially in text-attributed graph (TAG) tasks. Broadly, the role of language models in graph learning can be categorized into two main approaches: language models as enhancers and large language models as predictors~\cite{chen2024exploring}.

\subsubsection{Language models as enhancers}
Language models serve as enhancers by assisting graph models in representation learning and knowledge integration. Pre-trained language models (PLMs) like BERT~\cite{devlin2018bert}, DeBERTa~\cite{he2020deberta}, and XLNet~\cite{yang2019xlnet} are commonly used to transform raw textual descriptions into embeddings, improving graph models’ ability to capture node semantics. For instance, OFA~\cite{ofa} encodes text descriptions of nodes and edges into fixed-length vectors, unifying graph data from different domains and enabling strong performance in supervised, few-shot, and zero-shot settings. Similarly, GraphAlign~\cite{hou2024graphalign}, BooG~\cite{boog}, and ZeroG~\cite{li2024zerog} utilize PLMs to embed textual node features, ensuring feature consistency across diverse datasets during pre-training.

Beyond embedding textual features, large language models (LLMs) contribute to graph representation enrichment. TAPE~\cite{tape} generates textual explanations for model predictions, which are then transformed into additional node features, enhancing GNN-based learning. On the other hand, LLMGNN~\cite{llmgnn} uses LLMs to annotate a subset of nodes with high quality labels, which are then leveraged by GNNs to predict the remaining unlabeled nodes. This method effectively combines LLMs’ semantic reasoning with the structured learning power of GNNs. Building on this direction, Zhou et al.~\cite{zhou-etal-2025-graph} propose to treat each graph as a new language, translating structures into graph–language corpora to enable LLM pre-training that captures structural orders. GraphiT~\cite{khoshraftar2025graphit} further explores prompt optimization for efficient node classification on TAGs, while Zhang et al.~\cite{zhang2025rethinking} revisit graph structure learning under the LLM paradigm. For temporal settings, Zhang et al.~\cite{zhang2025cross} unify text semantics with graph structures for temporal TAGs, showing the versatility of LLMs across dynamic scenarios. Likewise, GRAIL~\cite{lim2025grail} investigates retrieval-augmented in-context learning, where node embeddings provide graph-aware contexts to LLMs, improving performance on real-world TAG benchmarks.

\subsubsection{Large language models as predictors}
LLMs can serve as direct predictors for graph-related tasks such as node classification and link prediction. Instruction tuning is a widely used technique to enhance LLMs’ predictive accuracy~\cite{ouyang2022training, sanh2021multitask}, helping them better interpret graph structures through task-specific prompts. For instance, InstructGLM~\cite{instructglm} employs multi-prompt tuning to integrate multi-hop structural information, improving its ability to capture complex relationships. GraphGPT~\cite{tang2024graphgpt} follows a dual-stage approach: first, it aligns structural information with language tokens via self-supervised graph matching, and second, it fine-tunes the model on task-specific instructions, leading to more accurate predictions.

Beyond standalone LLMs, hybrid models combine them with GNNs or graph transformers to better leverage graph structure. UniGraph~\cite{he2024unigraph} enhances zero-shot learning by aligning textual instructions with category labels while incorporating GNNs for structural learning. GraphLLM~\cite{chai2023graphllm} combines LLM with graph transformer to enrich LLM attention layers with structural and semantic information, enabling more effective graph reasoning. In contrast, LLaGA~\cite{chen2024llaga} avoids full LLM tuning and instead fine-tunes a lightweight graph projector, reducing computational cost while maintaining strong predictive performance. These approaches highlight the evolving role of LLMs in graph learning, demonstrating their flexibility in both direct prediction and hybrid architectures.

\section{Datasets}
\label{sec:Datasets}
We summarize the details of used datasets in Table \ref{tab:Datasets}. We convert all graphs into undirected graphs and remove self-loops.

\begin{table*}[htbp]
\centering
\small
% \scriptsize
\setlength{\tabcolsep}{0.7mm}

{

\begin{tabular}{@{}lcccccccc@{}}
\toprule
\rowcolor{gray!10}
\textbf{Dataset} & \textbf{Domain} & \textbf{Task}  & \textbf{\#Node} & \textbf{\#Edge} & \textbf{\#Classes} & \textbf{Metrics} & \textbf{Default feature}\\ \midrule
Cora   & citation & Node, Link      & 2,708     & 5,429 & 7 & Accuracy & Bag-of-Words \cite{nlgraph}\\
Pubmed  & citation & Node, Link      & 19,717    & 44,338 & 3 & Accuracy & TF-IDF\\
OGBN-Arxiv  & citation  & Node, Link      & 169,343   & 1,166,243 & 40 & Accuracy & Skip-gram \cite{skimgram}\\
OGBN-Products & e-commerce & Node, Link   & 2,449,029  & 61,859,140 & 47 & Accuracy & Bag-of-Words\\ 

Computer  & e-commerce & Node      & 87,229    & 721,081 & 10 & Accuracy & -\\
Reddit  & social network & Node      & 33,434    & 198,448 & 2 & Accuracy & -\\
Instagram  & social network & Node      & 11,339    & 144,010 & 2 & Accuracy & -\\
WikiCS  & web link & Node      & 11,701    & 215,863 & 10 & Accuracy & -\\
\bottomrule
\end{tabular}
\caption{Datasets}
\label{tab:Datasets}
}

\end{table*}

For Cora, PubMed, and OGBN-Arxiv, each node represents a paper and the edges denote co-citations. For OGBN-Products, nodes represent Amazon products and edges act as co-purchases. Due to the large size of OGBN-Products, we use Cluster-GCN \cite{clustergcn} to process it in smaller partitions. The structural information and label information of these datasets can be achieved from Pyg, and we will release the codes for raw feature processing. Below is some relevant information about each datasets:

\begin{itemize}[leftmargin=*]
    \item \textbf{Cora} \cite{cora}. Cora has seven categories: ["Rule Learning", "Neural Networks", "Case Based", "Genetic Algorithms", "Theory", "Reinforcement Learning", "Probabilistic Methods"]. The raw text attributes can be obtained from \footnote{\url{https://people.cs.umass.edu/mccallum/data.html}}.
    \item \textbf{PubMed} \cite{pubmed}. PubMed has three categories: ["Diabetes Mellitus, Experimental", "’Diabetes Mellitus Type 1", "Diabetes Mellitus Type 2"]. The raw text attributes can be obtained from TAPE \cite{tape} \footnote{\url{https://github.com/XiaoxinHe/TAPE}}.
    \item \textbf{OGBN-Arxiv and OGBN-Products} \cite{ogb}. OGB benchmark provides these two datasets. For OGBN-Arxiv, the raw text attributes can be downloaded from \footnote{\url{https://snap.stanford.edu/ogb/data/misc/ogbn-arxiv/titleabs.tsv.gz}}. For OGBN-Products, the raw text attributes can be downloaded from \footnote{\url{http://manikvarma.org/downloads/XC/XMLRepository.html}}.
\end{itemize}

In extended experiments, we use Computer, Reddit, Instagram, and WikiCS datasets. Computer is from E-Commerce Network, Reddit and Instagram are from Social Networks, and WikiCS represents web link network. We list the details below:
\begin{itemize}[leftmargin=*]
    \item \textbf{Computer}. Computer is from Amazon Electronics dataset \cite{ni2019justifying}, where each node represents an item in the Computer category. We use the processed dataset released in \cite{ofa}.
    \item \textbf{Reddit and Instagram}. A node represents a user, and edges denote whether two users have replied to each other. The raw text data is collected from \cite{huang2024can}.
    \item \textbf{WikiCS}. Each node represents a Wikipedia page, and each edge represents a reference link between pages. The raw text data is collected from \cite{ofa}.
\end{itemize}

\textbf{Data Split.} For node-level tasks, we use the standard train/validation/test splits \cite{ogb}: 6:2:2 for Cora, Pubmed, Computer, Reddit, Instagram, and WikiCS, 6:2:3 for the OGBN-Arxiv dataset ,and 8:2:90 for OGBN-Products. For link prediction, we randomly sample node pairs from the training nodes for training and from the testing nodes for evaluation. The size of the edge-level training set matches that of the node-level training set.

\section{Impacts of Different Node Feature Embedding Methods}
\label{sec:Impacts of Different Node Feature Embedding Methods}
Node features play a crucial role in node classification and link prediction tasks. For LLMs, raw text attributes are directly used as node features, while datasets like Cora, PubMed, Arxiv, and Products provide default preprocessed features generated through feature embedding methods (as shown in Table \ref{tab:Datasets}). This raises an important question: \textbf{is it fair to compare baseline models using default features with LLMs that rely on raw text attributes?}

To address this, we embedded the raw text attributes using various pre-trained LLMs and fed these embeddings into GraphSAGE for node classification tasks. The results are summarized in Table \ref{tab: Impacts of Different Node Feature Embedding Methods}. Specifically, all-MiniLM-L6-v2 is the latest Sentence-BERT model, and text-embedding-ada-002 is the latest embedding model from OpenAI.

From the results, we observe no significant accuracy improvements when using pre-trained LLM embeddings over the default node features. In some datasets, LLM-based embeddings perform better, while in others, default node features yield stronger results. Therefore, we believe that using the default node features provided by corresponding datasets is reasonable and fair.

\begin{table}[htbp]
\centering

\scalebox{0.64}{ % Adjust the scale as needed
\begin{tabular}{@{}lcccc@{}}
\toprule
\rowcolor{gray!10}
\textbf{Embedding Methods} & \textbf{Cora} & \textbf{PubMed} & \textbf{Arxiv} & \textbf{Products} \\ \midrule
default & 89.67 & 89.02 & 71.35 & 82.89 \\
all-MiniLM-L6-v2 \cite{sbert} & 89.88 & 89.91 & 72.03 & 81.82 \\
t5-small \cite{t5} & 86.71 & 87.78 & 70.28 & 79.64 \\
e5-base \cite{e5} & 88.10 & 87.12 & 71.52 & 80.33 \\
text-embedding-ada-002 & 89.30 & 89.72 & 72.20 & 82.45 \\

\bottomrule
\end{tabular}

}
\caption{Impacts of different node feature embedding methods. Task: node classification. Model: GraphSAGE}
\label{tab: Impacts of Different Node Feature Embedding Methods}
\end{table}

\section{Detailed Experimental Settings}
\label{sec:Detailed-Experimental-Settings}
\subsection{Computation Environment}
In this paper, all the experiments were conducted on one single server with 4 80G Nvidia A100 GPUs.

\subsection{Model Settings}
\begin{itemize}[leftmargin=*]

\item \textbf{GCN \& GraphSAGE}
\begin{lstlisting}[language=Python]
num_layers=3, hidden_channels=256, dropout=0.5, 
norm='batchnorm', activation='relu', 
optimizer=torch.optim.AdamW, lr=0.005, weight_decay=1e-4, 
scheduler=torch.optim.lr_scheduler.StepLR, step_size=20, gamma=0.5, 
patience=20, min_delta=1e-3, epochs=8000
\end{lstlisting}

\item \textbf{GAT}
\begin{lstlisting}
num_layers=3, hidden_channels=256, dropout=0.5, heads=2, 
norm='batchnorm', activation='relu', 
optimizer=torch.optim.AdamW, lr=0.005, weight_decay=1e-4, 
scheduler=torch.optim.lr_scheduler.StepLR, step_size=20, gamma=0.5, 
patience=20, min_delta=1e-3, epochs=8000
\end{lstlisting}

\item \textbf{MixHop}
\begin{lstlisting}
num_layers=2, hidden_channels=256, powers=[ [0,1,2], [0,1] ], dropout=0.6,
add_self_loops=True, activation='relu', aggregation='mixhop',
optimizer=torch.optim.AdamW, lr=0.005, weight_decay=1e-4,
scheduler=torch.optim.lr_scheduler.StepLR, step_size=20, gamma=0.5,
early_stopping=dict(patience=20, min_delta=1e-3), max_epochs=8000, log_interval=10
\end{lstlisting}

\item \textbf{GraphCL}
\begin{lstlisting}
Graph Encoder:
    -Backbone: GCN, -Hidden Channels: 128, -Activation: ReLU, -Optimizer: Adam, -lr=0.01, -Epochs: 100
Data Augmentations:
    -Feature Masking: mask_rate=0.3, -Edge Perturbation: perturb_rate=0.1
Contrastive Loss:
    -Normalization: L2 (dim=1), -Temperature: 0.5
Linear Classifier:
    -Input Features: 128, -Optimizer: Adam, -lr=0.01, -Epochs: 50 (supervised training)
\end{lstlisting}

\item \textbf{GraphMAE}
\begin{lstlisting}
Graph Encoder:
    -Backbone: GCN, -Hidden Channels: 256, -Activation: ReLU, -Optimizer: Adam, -lr=0.01, -Epochs: 200
Data Augmentations:
    -Feature Masking: mask_ratio=0.5 (encoder-level), -Random Masking: mask_rate=0.3 (training-level)
Reconstruction Loss:
    -Loss Function: MSE Loss, -Reconstruction Target: Masked node features
Linear Classifier:
    -Input Features: 256, -Optimizer: Adam, -lr=0.01, -Epochs: 100 (supervised training)
\end{lstlisting}

\item \textbf{Graphormer}
We follow the hyper-parameter settings in the original paper \cite{graphormer}.

\item \textbf{Prodigy}
We follow the hyper-parameter settings in the original paper \cite{huang2024prodigy}.

\item \textbf{OFA}
We follow the hyper-parameter settings in the original paper \cite{ofa}.

\item \textbf{GIANT \& TAPE}
\begin{lstlisting}
gnn_type='GraphSAGE', num_layers= [2, 3, 4], hidden_channels= [128, 256],
optimizer=torch.optim.Adam, lr=0.001, weight_decay=0, dropout= [0.3, 0.5, 0.6] 
\end{lstlisting}

\item \textbf{All in one \& GPF-plus \& GraphPrompt}
\begin{lstlisting}
gnn_type='GCN', num_layers=2, hidden_channels=128, JK='last',
prompt_type=['All in one', 'GPF-plus', 'GraphPrompt'],
optimizer=torch.optim.Adam, lr=0.001, weight_decay=0, dropout=0.5,
epochs=800, batch_size=128, shot_num=5,
\end{lstlisting}

For detailed prompt designs, we follow the original papers \cite{sun2023allinone}, \cite{gpf-plus}, and \cite{liu2023graphprompt}.

\item \textbf{ZeroG}
We follow the hyper-parameter settings in the original paper \cite{li2024zerog}.

\item \textbf{LLaGA}
We follow the hyper-parameter settings in the original paper \cite{chen2024llaga}.

\item \textbf{Llama3B \& Llama8B}
\begin{lstlisting}
LLM Configuration: 
    -Base Model: [meta-llama/Llama-3.2-3B-Instruct, meta-llama/Llama-3.1-8B-Instruct], 
    -Use LoRA: true, -Max Sequence Length: 1024, -Model Precision: bfloat16
LoRA Configuration:
    -LoRA Rank (r): 16, -LoRA Alpha: 32, -LoRA Dropout: 0.05, 
    -Target Modules: [o_proj, gate_proj, down_proj, up_proj]
Training Configuration:
    -Optimizer: adamw_torch, -Learning Rate: 4e-4, -Train Batch Size: 2 x 12 (per_device x grad_accum), 
    -Total Epochs: 1, -Gradient Accu Steps: 12, -Pad Token ID: -100 (IGNORE_INDEX)
DeepSpeed Optimization:
    -Zero Stage: 2, -Offload Strategy: [-Optimizer -> CPU (pinned) ,-Activation Checkpointing: true], 
    -Pipeline Parallel: [-Enabled: true, -Micro Batch Size: 1]
Data Processing:
    -Data Sources: [Cora, PubMed, Arxiv, Products], -Input Format: System Prompt + User Query + Answer, 
    -Data Limits: [-Product/node: max 3,000 samples, -Product/link: max 2,000 samples], 
    -Preprocessing Workers: 20,
    -Cora & PubMed & Arxiv: [-Max 1-hop neighbors: 20, -Max 2-hop neighbors: 5],
    -Products: [-Max 1-hop neighbors: 30, -Max 2-hop neighbors: 10]
    
\end{lstlisting}

\end{itemize}

\section{Hyperparameter Search Space}
\label{sec:appendix_hyperparameters}

For transparency and reproducibility, this section details the hyperparameter search spaces explored during the tuning of our baseline models. The final parameters selected for the experiments are reported in Appendix \ref{sec:Detailed-Experimental-Settings}. All selections were made based on the best-performing validation accuracy.

\subsection{GNN Models (GCN, GraphSAGE, GAT)}
For the graph neural network baselines, we performed a grid search over the following hyperparameters on each dataset:
\begin{itemize}[leftmargin=*]
    \item \textbf{Learning Rate}: \{0.01, 0.005, 0.001\}
    \item \textbf{Hidden Channels}: \{128, 256\}
    \item \textbf{Number of Layers}: \{2, 3\}
    \item \textbf{Dropout Rate}: \{0.3, 0.5, 0.6\}
    \item \textbf{Weight Decay}: \{5e-4, 1e-4, 5e-5\}
\end{itemize}

\subsection{LoRA Fine-Tuning for Llama Models}
For the instruction tuning of Llama3B and Llama8B using Low-Rank Adaptation (LoRA), we explored a focused set of hyperparameters known to be effective for this technique. The primary goal was to find a stable configuration without conducting an exhaustive, computationally prohibitive search.
\begin{itemize}[leftmargin=*]
    \item \textbf{Learning Rate}: \{4e-4, 2e-4, 1e-4\}
    \item \textbf{LoRA Rank (r)}: \{8, 16, 32\}
    \item \textbf{LoRA Alpha ($\alpha$)}: \{16, 32\}. We maintained the common practice of setting alpha to be twice the rank.
    \item \textbf{LoRA Dropout}: \{0.05, 0.1\}
\end{itemize}
The target modules for LoRA (o\_proj, gate\_proj, etc.) were kept consistent across all runs as they are standard choices for Llama-family models. The final chosen configuration was a learning rate of 4e-4, a rank of 16, an alpha of 32, and dropout of 0.05, which provided a good balance of performance and stability.

\section{Baseline models}
\label{sec:Baseline models}
In this paper, we evaluate multiple baseline models and provide detailed descriptions of their implementations as follows. These models were applied to a consistently preprocessed version of the datasets to ensure fair comparisons and produce the experimental results presented in this study.
\begin{enumerate}[leftmargin=*]
    \item \textbf{GNNs}: For GCN \cite{gcn}, GraphSAGE \cite{graphsage}, GAT \cite{gat}, and Mixhop \cite{abu2019mixhop}, we follow the models on OGB Leaderboards \footnote{\url{https://ogb.stanford.edu/docs/leader_nodeprop/}}. Specifically, the first three models are all from \cite{luo2024classic}, and the codes can be obtained from \footnote{\url{https://github.com/LUOyk1999/tunedGNN}}.
    
    \item \textbf{Graph SSL Models}: We choose GraphCL \cite{you2020graphcl} and GraphMAE \cite{hou2022graphmae} in this categories. GraphCL employs contrastive learning by distinguishing augmented views of the same graph from others, while GraphMAE uses masked autoencoding, reconstructing masked graph components to learn node representations without requiring augmented views. For GraphCL, we follow the implementation from \footnote{\url{https://github.com/Shen-Lab/GraphCL}}. For GraphMAE, we follow the implementation from \footnote{\url{https://github.com/THUDM/GraphMAE}}.
    
    \item \textbf{Graph Transformers}: We use Graphormer \cite{graphormer} in this categories. Graphormer is a transformer-based model designed specifically to handle graph-structured data, enabling efficient processing and analysis of complex relational information.The implementation is from \footnote{\url{https://github.com/microsoft/Graphormer}}.

    \item \textbf{Foundational Graph Prompt Models}: We use Prodigy \cite{huang2024prodigy}, OFA \cite{ofa}, All in one \cite{sun2023allinone}, GPF-plus \cite{gpf-plus}, GraphPrompt \cite{liu2023graphprompt}, and ZeroG \cite{li2024zerog} in this categories. 
    \begin{itemize}[leftmargin=*]
        \item Prodigy enables in-context learning over graphs by utilizing a novel prompt graph representation and a family of in-context pretraining objectives, achieving superior performance on diverse downstream classification tasks without the need for retraining.

        \item OFA represents nodes and edges as human-readable text, mapping them from various domains into a unified space using LLMs. The framework then adapts to different tasks by embedding task-specific prompts within the input graph.

        \item All in one proposes a novel method to unify graph prompts and language prompts, enhancing the performance of various graph tasks through effective prompt design and meta-learning techniques.

        \item GPF-plus is an enhanced graph prompt tuning method that assigns independent learnable vectors to each node, offering great flexibility and expressiveness and consistently outperforming other methods in various experiments.

        \item GraphPrompt leverages a common task template based on subgraph similarity, enhanced with task-specific learnable prompts to improve performance across different tasks such as node and graph classification.

        \item ZeroG uses a language model to encode node features and class labels, incorporating prompt-based subgraph sampling and efficient fine-tuning techniques to tackle the challenges of cross-dataset zero-shot transferability in graph learning.
    \end{itemize}
     The implementations of Prodigy and OFA can be obtained from \footnote{\url{https://github.com/snap-stanford/prodigy}} and \footnote{\url{https://github.com/LechengKong/OneForAll}}, respectively. For All in one, GPF-plus, and GraphPrompt, we use the implementation from ProG \cite{zi2024prog} \footnote{\url{https://github.com/sheldonresearch/ProG}}. For ZeroG, we follow the implementation from \footnote{\url{https://github.com/NineAbyss/ZeroG}}.

    \item \textbf{LM-Augmented Graph Learning Models}: We choose GIANT \cite{giant} and TAPE \cite{tape}. GIANT conducts neighborhood prediction using XR-Transformers \cite{zhang2021fast}, resulting in an LLM that generates superior feature vectors for node classification compared to traditional bag-of-words and standard BERT embeddings. TAPE uses explanations from LLMs as features to enhance the performance of GNNs on text-attributed graphs, achieving state-of-the-art results on various benchmarks with significantly lower computation time. For GIANT and TAPE, we follow the implementation from \footnote{\url{https://github.com/NineAbyss/GLBench}}.

    \item \textbf{LLM with Graph Projectors}: LLaGA \cite{chen2024llaga} is chosen for this category. The implementation is from \footnote{\url{https://github.com/VITA-Group/LLaGA}}.
    
\end{enumerate}

\section{Theoretical Justification: How LLMs Emulate Graph Reasoning}
\label{app:theoretical_justification}

A primary concern regarding the application of LLMs to graph tasks is the fundamental modality mismatch: LLMs are pre-trained on sequential text, whereas graphs are inherently non-sequential, relational data structures. This section provides a theoretical and mechanistic justification for why LLMs, particularly through the mechanisms of prompting and instruction tuning, can successfully perform graph reasoning. Our core argument is that \textbf{the Transformer's self-attention mechanism can be viewed as a powerful and general form of a graph operator, which learns to emulate the message-passing operations of Graph Neural Networks (GNNs) when guided by structured textual prompts and task-specific fine-tuning.}

\subsection{The Self-Attention Mechanism as a General Graph Operator}
The Transformer architecture is fundamentally a relational reasoner. For a sequence of $n$ token embeddings $X \in \mathbb{R}^{n \times d}$, the self-attention mechanism computes a new set of representations $Z \in \mathbb{R}^{n \times d}$ as:
\begin{equation}{
Z = \text{Attention}(Q, K, V) = \underbrace{\text{softmax}\left(\frac{QK^T}{\sqrt{d_k}}\right)}_{A}V
}
\end{equation}
The attention matrix $A \in \mathbb{R}^{n \times n}$ contains pairwise scores $A_{ij}$ representing the influence of token $j$ on token $i$. This allows us to interpret the attention mechanism as performing an update on a \textbf{dynamic, fully-connected, weighted graph} $\mathcal{G}_A = (\mathcal{T}, A)$, where the set of tokens $\mathcal{T}$ are the nodes.

The representation $z_i$ for the $i$-th token is an aggregation over all tokens in the sequence:
\begin{equation}{
z_i = \sum_{j=1}^{n} A_{ij} v_j
}
\end{equation}
This is a generalized form of neighborhood aggregation, where the "neighborhood" of each token is the entire sequence, and the weights are learned based on context.

\subsection{Emulating GNN Message Passing via Prompt-Structured Attention}
The success of GNNs stems from the message-passing paradigm. For a node $v$ in a graph $\mathcal{G} = (\mathcal{V}, \mathcal{E})$, the update rule for its hidden representation $h_v^{(l)}$ at layer $l$ is:
\begin{align}
    m_{\mathcal{N}(v)}^{(l+1)} &= \text{AGGREGATE}^{(l)}\left(\{h_u^{(l)} : u \in \mathcal{N}(v)\}\right) \label{eq:gnn_agg} \\
    h_v^{(l+1)} &= \text{UPDATE}^{(l)}\left(h_v^{(l)}, m_{\mathcal{N}(v)}^{(l+1)}\right) \label{eq:gnn_update}
\end{align}
We posit that an LLM, when given a structured textual prompt, emulates this two-step process. Let the input prompt serialize a target node $v$ and its neighborhood $\mathcal{N}(v)$. The set of all tokens $\mathcal{T}$ can be partitioned into three disjoint sets: tokens representing the target node itself ($\mathcal{T}_v$), tokens representing its neighbors ($\mathcal{T}_{\mathcal{N}(v)}$), and all other tokens ($\mathcal{T}_{\text{other}}$).

The attention-based update for a token $i \in \mathcal{T}_v$ can then be decomposed as:
\begin{equation}
\resizebox{\columnwidth}{!}{
$z_i = \underbrace{\sum_{j \in \mathcal{T}_v} A_{ij} v_j}_{\text{Self-update}} + \underbrace{\sum_{j \in \mathcal{T}_{\mathcal{N}(v)}} A_{ij} v_j}_{\text{Neighbor Aggregation}} + \underbrace{\sum_{j \in \mathcal{T}_{\text{other}}} A_{ij} v_j}_{\text{Contextual Noise/Signal}}$
}
\end{equation}
This decomposition reveals the analogy to GNNs:
\begin{itemize}[leftmargin=*]
    \item \textbf{Aggregation:} The term $\sum_{j \in \mathcal{T}_{\mathcal{N}(v)}} A_{ij} v_j$ is a direct analogue to the GNN \textbf{AGGREGATE} function (Eq. \ref{eq:gnn_agg}). The LLM learns to assign high attention scores $A_{ij}$ to tokens representing the true neighbors of node $v$, effectively aggregating their information.
    \item \textbf{Update:} The aggregated message is then combined with the node's own representation (the self-update term) and passed through a position-wise Feed-Forward Network (FFN):
    \begin{equation}{
    h'_i = \text{FFN}(\text{LayerNorm}(z_i + x_i))
    }
    \end{equation}
    This FFN, a powerful non-linear transformer, serves as the \textbf{UPDATE} function (Eq.\ref{eq:gnn_update}), producing the final, contextually-aware representation for the token.
\end{itemize}
By providing $k$-hop neighbor information, we allow the LLM to implicitly simulate a $k$-layer GNN within a single forward pass.

\subsection{Learning Graph Inductive Biases via Instruction Tuning}
An off-the-shelf LLM, while architecturally capable, lacks the specific \textit{inductive biases} for graph structures. It has no inherent reason to prioritize tokens labeled "neighbor" over any other tokens. \textbf{Instruction tuning} instills these biases by optimizing the model's parameters $\theta$ on a graph-specific objective.

Let $\mathcal{D}_{\text{graph}} = \{(\text{prompt}(G_k, v_k), y_k)\}_{k=1}^M$ be a dataset of graph-based instruction-response pairs. The tuning process minimizes a loss function, typically the cross-entropy for classification tasks:
\begin{equation}{
\mathcal{L}_{\text{tune}}(\theta) = -\sum_{k=1}^M \log P(y_k | \text{prompt}(G_k, v_k); \theta)
}
\end{equation}
The key effect of minimizing $\mathcal{L}_{\text{tune}}$ is the reshaping of the attention matrix $A$. The optimization process implicitly forces the attention patterns to reflect the graph structure. Specifically, for a target node token $i$ and a neighbor token $j$, the fine-tuning process encourages:

\begin{equation}
\resizebox{\columnwidth}{!}{
$\theta^* = \arg\min_\theta \mathcal{L}_{\text{tune}}(\theta) \implies A_{ij} \text{ is high if token } j \in \mathcal{T}_{\mathcal{N}(v_i)}$
}
\end{equation}
Essentially, instruction tuning teaches the LLM that for graph-related prompts, the "correct" reasoning path involves focusing attention on the explicitly provided neighborhood information. It transforms the generic, semantically-driven attention into a specialized, structurally-aware attention mechanism that mimics the sparse connectivity of the original graph.

\subsection{Synthesis: When and Why LLMs Succeed or Falter}
This theoretical framework helps explain our empirical findings:
\begin{itemize}[leftmargin=*]
    \item \textbf{Success in Local Tasks:} LLMs excel at tasks like node classification and link prediction because these primarily rely on local structures ($1$-hop, $2$-hop), which can be effectively encoded and processed by the partitioned attention mechanism described above.
    \item \textbf{Impact of Instruction Tuning:} The significant performance gap between pre-trained and tuned LLMs is explained by the optimization of $\mathcal{L}_{\text{tune}}$, which is necessary to instill graph-centric inductive biases into the model's attention patterns.
    \item \textbf{Potential Limitations:} This framework also predicts limitations. Tasks requiring \textbf{global graph properties} (e.g., diameter, global clustering coefficient) are challenging because the full structure, and thus the complete attention graph, cannot fit into a finite context window. Furthermore, for purely algorithmic tasks (e.g., shortest path), LLMs act as probabilistic pattern matchers rather than deterministic solvers, leading to approximations. Classical algorithms, which operate on the true graph adjacency matrix, remain more precise and efficient for such problems.
\end{itemize}
In conclusion, the success of LLMs in graph tasks is a direct consequence of the Transformer architecture's inherent ability to model relational data. This latent ability is unlocked and specialized through instruction tuning, which aligns the self-attention mechanism to emulate the message-passing framework of GNNs.

\section{In-depth Analysis of LLM Sensitivity to Prompt Variations}
\label{sec:Analysis of LLM Sensitivity}

\subsection{Motivation}
A critical observation from our main experiments is that the performance of LLMs on graph learning tasks is highly sensitive to the prompt format. To address the reviewer's feedback for a deeper understanding of this phenomenon, this section presents a systematic investigation into \textit{why} this sensitivity exists. We aim to dissect the components of a prompt and analyze their individual impact on the model's reasoning process for node classification.

\subsection{Experimental Design}
To isolate the factors influencing performance, we conducted a controlled experiment with the following setup:
\begin{itemize}[leftmargin=*]
    \item \textbf{Model:} We use \textbf{Llama8B} as our representative model.
    \item \textbf{Task and Dataset:} We focus on the \textbf{node classification} task on the \textbf{Cora} dataset, allowing for a focused and granular analysis.
    \item \textbf{Methodology:} We establish a \textbf{baseline prompt} using the exact performance reported in our main experiments (Table \ref{tab:node_classification_results_LLM}). We then introduce a series of controlled, single-factor variations. By modifying only one aspect of the prompt at a time, we attribute any resulting performance change directly to that modification.
\end{itemize}

We designed three categories of prompt variations to test distinct hypotheses about LLM sensitivity:
\begin{enumerate}[leftmargin=*]
    \item \textbf{Instruction Verbosity:} How does the level of detail in the task description affect the model's focus and understanding?
    \item \textbf{Structural Phrasing:} How crucial are explicit graph-related terms (e.g., "neighbor") compared to more natural, relational phrasing?
    \item \textbf{Task Command Wording:} How robust is the model to semantic but trivial paraphrasing of the core instruction?
\end{enumerate}

\subsection{Results and Analysis}
The performance of Llama-8B under these controlled prompt variations is summarized in Table \ref{tab:prompt_sensitivity}. Our analysis of these results reveals several key insights into why prompt format is so influential:

\begin{table*}[h!]
\centering
\small

\begin{tabular}{@{}llc@{}}
\toprule
\textbf{Variation Category} & \textbf{Prompt Description} & \textbf{Accuracy (\%)} \\ \midrule
\textbf{Baseline} & \textbf{Standard "1-hop w/o label" prompt} & \textbf{58.4} \\ \midrule
\multirow{2}{*}{Instruction Verbosity} & \textit{Minimal Instruction}: "Given the target and neighbors, predict its category." & 55.5 \\
 & \textit{Verbose Instruction}: "You are an expert paper classifier..." & 56.8 \\ \midrule
\multirow{2}{*}{Structural Phrasing} & \textit{Relational}: "...target paper is connected to the following papers..." & 55.1 \\
 & \textit{Implicit}: "Target Paper: ... Related Papers: ..." & 53.7 \\ \midrule
\multirow{2}{*}{Task Command Wording} & \textit{Synonym 1}: "...assign the correct label to the Target node." & 58.9 \\
 & \textit{Synonym 2}: "What is the research area of the Target node?" & 58.1 \\ \bottomrule
\end{tabular}
\caption{Performance analysis of Llama-8B on the Cora node classification task under different prompt variations. The baseline is the standard "1-hop w/o label" prompt.}
\label{tab:prompt_sensitivity}
\end{table*}

\paragraph{1. Clarity and Sufficient Context are Crucial.}
The \textit{Minimal Instruction} prompt, which strips away the guiding context (e.g., "You are a good graph reasoner..."), causes a performance drop of 3 percentage points. This suggests that LLMs benefit from "role-playing" or context-setting instructions that frame the problem. Without this frame, the model may struggle to activate the most relevant reasoning pathways. Conversely, the \textit{Verbose Instruction} prompt, while providing more detail, slightly underperforms the baseline. This indicates a trade-off: while some context is essential, excessive detail may introduce irrelevant information that dilutes the focus on the core data, acting as noise. The standard baseline prompt appears to strike an effective balance.

\paragraph{2. Explicit Structural Language Bridges the Gap between Text and Graphs.}
The most significant finding comes from the \textit{Structural Phrasing} variations. Replacing the explicit, technical term "Known neighbor papers at hop 1" with more natural language like "connected to" (\textit{Relational}) or simply "Related Papers" (\textit{Implicit}) consistently degrades performance. The accuracy drop is most severe with the implicit phrasing. This strongly suggests that LLMs, being pre-trained on sequential text, do not inherently interpret a list of items following a target as a formal graph structure. Explicit keywords like \textbf{"hop"} and \textbf{"neighbor"} act as powerful signals that help the model shift from a standard text-processing mode to a graph-reasoning mode. These terms provide an unambiguous structural scaffold that is otherwise missing, forcing the model to recognize and leverage the relational nature of the input data.

\paragraph{3. LLMs are Robust to Simple Paraphrasing of the Core Task.}
The \textit{Task Command Wording} variations show almost no change in performance. Replacing "predict the category" with "assign the label" or phrasing it as a question ("What is the research area...") results in nearly identical accuracy. This demonstrates that the model has a robust semantic understanding of the core objective. Its sensitivity is not rooted in surface-level vocabulary for the main task command, but rather in the description and framing of the input data's structure.

\section{Extended Experiments}
\label{sec:Extended Experimental Results}

\subsection{LLM Understanding of Graph Structures}
\label{sec:Case 5}
The structure of a graph sets it apart from natural language, and the model ability to comprehend these structures is vital for enhancing its performance on graph tasks. In this section, we explore the ability of instruction-tuned LLMs to understand graph structures.

\subsubsection{Experiment Settings}
We remove all node attributes and retain only node IDs to eliminate the influence of attributes on LLM reasoning. Examples of these prompt formats are provided in Appendix \ref{sec:Prompt Formats for Pure Graph Structure}.

\begin{figure}[]
  \centering
  \scriptsize
  \includegraphics[width=1\linewidth]{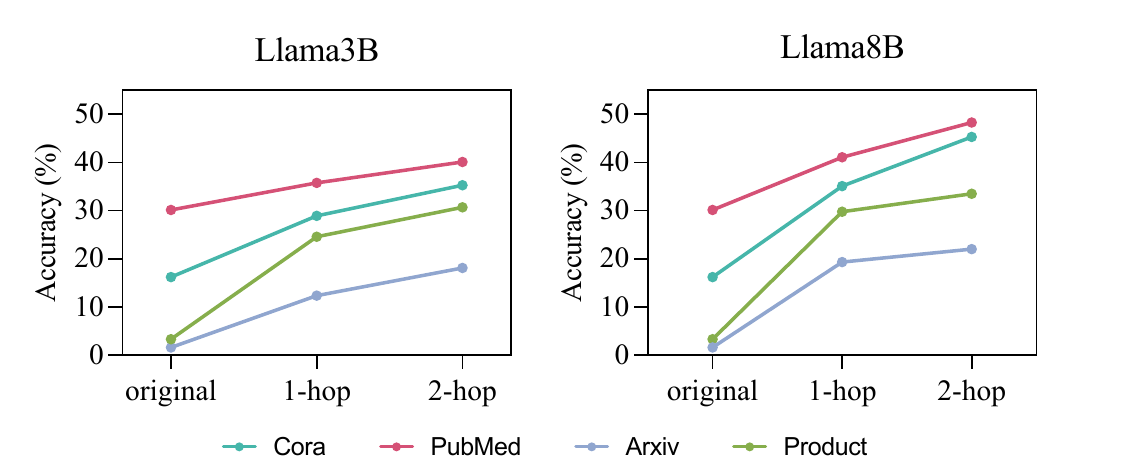}
  %\vspace{-20pt}
  \caption{LLM performance on node classification without node attributes}
  \label{fig:LLM performance on node classification without node attributes}
\end{figure}

\subsubsection{Results}
\paragraph{\underline{Node classification}}
We present the results of node classification in Figure \ref{fig:LLM performance on node classification without node attributes}. “Original” refers to Llama3B or Llama8B without parameter optimization, while “1-hop” and “2-hop” correspond to 1-hop w/o label and 2-hop w/o label, respectively. From the figure, we observe that off-the-shelf LLMs perform similarly to random guessing in node classification. For instance, with 7 classes in Cora, the probability of random guessing correctly is 14.28\%, and the experimental results align closely with this probability. This is because LLMs struggle to make accurate predictions based purely on graph structure without semantic information. After instruction tuning, LLMs start to learn some graph structural information, leading to improved accuracy. However, the improvement is limited, likely because the classes in these datasets are strongly correlated with node features, and the graph structural differences between categories are minimal. This explains why simpler models like MLPs \cite{hu2021graph} and our ego prompt format perform relatively well, as they rely more on the node features than on the graph structure itself.

\begin{table}[htbp]
\centering
\small
\scriptsize
\setlength{\tabcolsep}{0.2mm}

{%
\begin{tabular}{l c c c c c c}
\toprule
\rowcolor{gray!10}
\textbf{Models} & \textbf{Prompts} & \textbf{Cora} & \textbf{PubMed} & \textbf{ArXiv} & \textbf{Products} & \textbf{Avg} \\ 
\midrule
\multirow{2}{*}{Llama3B w attributes} & 1-hop & 72.97 & 71.55 & 72.45 & 78.92 & 73.97 \\
& 2-hop & 68.21 & 59.95 & 68.55 & 79.17 & 68.97 \\
\midrule

\multirow{2}{*}{Llama8B w attributes} & 1-hop & 80.44 & 74.80 & 87.80 & 85.29 & 82.08 \\
& 2-hop & 89.39  & 77.30 & \underline{92.30}  & 90.77  & 87.44  \\
\midrule
\multirow{2}{*}{Llama3B w/o attributes} & 1-hop & 66.61 & 55.44 & 64.94 & 78.47 & 66.37 \\
& 2-hop & 72.22 & 58.62 & 65.62 & 74.52 & 67.75 \\
\midrule
\multirow{2}{*}{Llama8B w/o attributes} & 1-hop & 63.19 & 55.81 & 68.62 & 81.30 & 67.23 \\
& 2-hop & 85.58 & 69.50 & 84.88 & 87.78 & 81.94 \\
\midrule
\multirow{2}{*}{tuned Llama3B w/o attributes} & 1-hop & 75.88 & 74.70 & 78.30 & 77.38 & 76.57 \\
& 2-hop & \underline{93.20}  & \textbf{97.66}  & 89.00 & \underline{94.09}  & \underline{93.49}  \\
\midrule
\multirow{2}{*}{tuned Llama8B w/o attributes} & 1-hop & 85.15 & 78.81  & 89.34  & 87.98 & 85.32 \\
& 2-hop & \textbf{94.11}  & \underline{97.44}  & \textbf{93.67}  & \textbf{94.54}  & \textbf{94.94}  \\
\bottomrule
\end{tabular}
\caption{LLM performance on link prediction without node attributes. Llama3B w attributes and Llama8B w attributes are for comparison. The \textbf{best} and \underline{second-best} are highlighted.}
\label{tab:LLM performance on link prediction without node attributes}
}

\end{table}

\paragraph{\underline{Link prediction}}
From Table \ref{tab:LLM performance on link prediction without node attributes}, we observe that LLMs with node attributes outperform those without, highlighting the positive role of node attributes in LLM reasoning. However, after instruction tuning without node attributes, the LLMs show a significant improvement in link prediction accuracy. This demonstrates that LLMs can effectively learn and understand graph structures, achieving high link prediction accuracy even in the absence of node attributes.

\begin{remark}
LLMs can learn graph structures effectively through instruction tuning. While node attributes improve performance, LLMs can still achieve high accuracy in link prediction by leveraging structural information alone. However, the improvement in node classification is limited, likely because the classes are closely related to node features and the structural differences between categories are minimal.
\end{remark}

\subsection{Robustness of LLMs}
\label{sec:Robustness of LLMs}
We aim to investigate the robustness of LLMs under two challenging conditions: missing edge information and decreasing graph homophily. Graph homophily refers to the tendency of similar nodes to connect. Our goal is to understand whether LLMs primarily rely on node similarity when performing graph reasoning and how reducing this similarity affects their performance.

\subsubsection{Experiment Settings}
We conduct experiments on the \textbf{Cora} and \textbf{ArXiv} datasets, designing two scenarios: \textbf{drop same} and \textbf{drop random}. The former examines how reducing node similarity affects LLM performance, while the latter investigates the impact of simply reducing the number of edges.

\begin{itemize}[leftmargin=*]
    \item \textbf{Drop Same}: We randomly remove \textbf{0\%, 20\%, 40\%, 60\%, 80\%, and 100\%} of edges connecting nodes of the same class. This reduces node similarity, effectively lowering the homophily ratio \cite{loveland2024performance, whenandwhy}.
    \item \textbf{Drop Random}: We randomly remove edges but have to ensure that the number of dropped edges matches the corresponding \textbf{"drop same"} setting. For example, if 40\% "drop same" results in 1,000 removed edges, then 40\% "drop random" also removes 1,000 edges.
\end{itemize}

For LLMs, we use \textbf{DeepSeek V3} and \textbf{Llama3B}. As baselines, we include \textbf{GCN}, \textbf{GraphSAGE}, and \textbf{MixHop} \cite{abu2019mixhop} (which performs well on heterophilic graphs). All trainable models (GCN, GraphSAGE, MixHop, and Llama3B) are trained on graphs with varying levels of edge removal. Specifically, for each dataset (Cora and ArXiv), we train \textbf{twelve} models per method—six under "drop same" and six under "drop random", corresponding to the six drop percentages.

\begin{figure*}[htbp]
  \centering
  \includegraphics[width=1\linewidth]{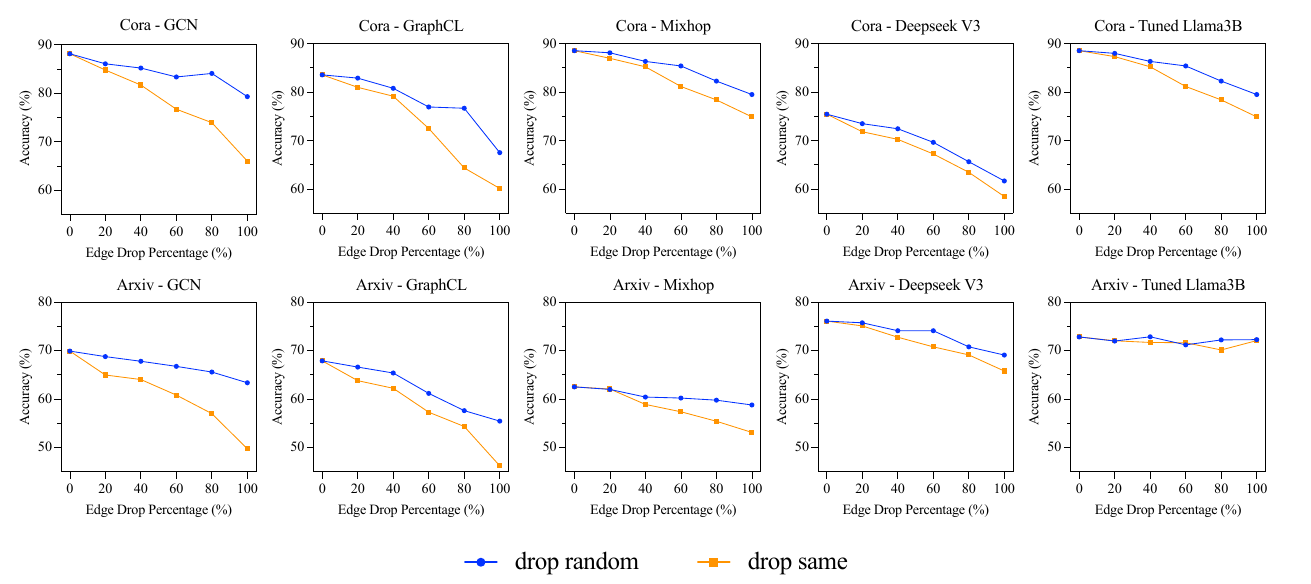}
  %\vspace{-20pt}
  \caption{Robustness of LLMs}
  \label{fig:Robustness of LLMs}
\end{figure*}

\subsubsection{Results}
We summarize the experimental results in Figure \ref{fig:Robustness of LLMs}. As expected, accuracy declines across all models and datasets as the edge drop percentage increases. However, the impact of edge removal is not uniform. The "drop same" condition leads to a sharper decline compared to "drop random", suggesting that reducing node similarity (homophily) has a greater negative effect than simply removing edges at random.

Interestingly, DeepSeek V3 and tuned Llama3B show more resilience to homophily reduction compared to GCN, GraphCL, and even Mixhop, indicating that they rely less on node similarity for classification. Among them, tuned Llama3B stands out, not only preserving high accuracy despite edge removal but also showing the lowest dependency on node similarity. This highlights that instruction tuning significantly enhances the robustness of LLMs, making them more adaptable to structural perturbations.

\begin{remark}
Reducing homophily (via “drop same”) has a more significant negative impact than randomly removing edges. LLMs, especially those after instruction tuning are more resilient to structural perturbations compared to GNNs like GCN, GraphSAGE, and even MixHop.
\end{remark}

\subsection{Computational Overhead Analysis}
\label{sec:Computational Overhead Analysis}

Computational overhead is an important consideration for real-world deployment. We evaluate both the training and inference times of several baseline models and LLMs with instruction tuning. The results are presented in Table \ref{tab:Training times} and Table \ref{tab:Inference times}. All measurements were conducted on a single NVIDIA A100-80G GPU.

Based on the results, we observe that during training, graph-specific models incur significantly lower computational overhead compared to LLM-based methods. For instance, the training time of Llama8B exceeds that of classic GNNs by more than 100×. This highlights the importance of LLM transferability (discussed in Section 5.3: if a one-time training process can support multiple downstream tasks, the high training cost may be justified. Therefore, a promising research direction is to further improve the adaptability and generalization of LLMs across different graph domains and tasks.

As for inference, although LLM-based models still require more time than graph-specific ones, the difference is less critical since all inference times are within the millisecond range. Thus, unless strict real-time performance is required, the overhead gap is relatively negligible.

\begin{table*}[htbp]
\centering
\small

{%
\begin{tabular}{l c c c c | c c c c}
\toprule
\rowcolor{gray!10}
 & \multicolumn{4}{c}{\textbf{Node classification}} & \multicolumn{4}{c}{\textbf{Link prediction}} \\
\cline{2-5} \cline{6-9}
\rowcolor{gray!10}
\textbf{Models} & \textbf{Cora} & \textbf{PubMed} & \textbf{ArXiv} & \textbf{Products} & \textbf{Cora} & \textbf{PubMed} & \textbf{ArXiv} & \textbf{Products}  \\ 
\midrule

GCN & 15.9s & 34.9s & 8.4m & 15m & 9.5s & 26.8s & 7.7m & 13.4m \\
GraphSAGE & 14.1s & 33.3s & 7.8m & 13.8m & 8.2s & 23.7s & 7.2m & 12m \\
GAT & 20.5s & 45s & 10.1m & 18.4m & 10.8s & 32.9s & 8.1m & 15.2m \\
GraphCL & 2.1m & 4.1m & 53m & 1.2h & 2m & 3.2m & 48.2m & 1.1h \\
GraphMAE & 3.8m & 6.3m & 1.1h & 1.5h & 3.2m & 4.9m & 1h & 1.4h \\
LLaGA & 13.8m & 29m & 6h & 8.4h & 12m & 27.5m & 5.2h & 7.9h \\
\midrule
Llama3B & 1.2h & 1.9h & 18.3h & 23.9h & 1.7h & 2.3h & 23.2h & 26.9h \\
Llama8B & 1.8h & 2.6h & 25.7h & 31.1h & 2.1h & 3h & 30h & 35.8h \\
\bottomrule
\end{tabular}
\caption{Training times of different models on node classification and link prediction tasks. We use 9 prompt formats to train LLMs on link prediction. LLM tuning was done on 4 A100-80G GPUs, so all reported times are multiplied by 4.}
\label{tab:Training times}
}

\end{table*}

\begin{table*}[htbp]
\centering
\small
% \scriptsize
% \setlength{\tabcolsep}{1mm}

{%
\begin{tabular}{l c c c c | c c c c}
\toprule
\rowcolor{gray!10}
 & \multicolumn{4}{c}{\textbf{Node classification}} & \multicolumn{4}{c}{\textbf{Link prediction}} \\
\cline{2-5} \cline{6-9}
\rowcolor{gray!10}
\textbf{Models} & \textbf{Cora} & \textbf{PubMed} & \textbf{ArXiv} & \textbf{Products} & \textbf{Cora} & \textbf{PubMed} & \textbf{ArXiv} & \textbf{Products}  \\ 
\midrule

GCN & 8ms & 12ms & 33ms & 40ms & 3ms & 5ms & 26ms & 38ms \\
GraphSAGE & 12ms & 29ms & 35ms & 3ms & 4ms & 24ms & 27ms & 33ms \\
GAT & 8ms & 10ms & 35ms & 38ms & 4ms & 5ms & 29ms & 41ms \\
GraphCL & 12ms & 19ms & 69ms & 71ms & 7ms & 10ms & 50ms & 71ms \\
GraphMAE & 15ms & 22ms & 76ms & 80ms & 8ms & 11ms & 57ms &  69ms\\
LLaGA & 40ms & 69ms & 112ms & 159ms & 27ms & 37ms & 99ms & 134ms \\
\midrule
Llama3B & 186ms & 211ms & 338ms & 401ms & 113ms & 139ms & 172ms & 228ms \\
Llama8B & 231ms & 238ms & 381ms & 459ms & 127ms & 151ms & 203ms & 273ms \\
\bottomrule
\end{tabular}
\caption{Inference times of different models on node classification and link prediction tasks.}
\label{tab:Inference times}
}

\end{table*}

\subsection{Comparison of Different LLMs on Node Classification}
\label{sec:Comparison of Different LLMs on Node Classification}

\begin{table}[htbp]
\centering
\small
\renewcommand{\arraystretch}{0.9}
\scriptsize
\setlength{\tabcolsep}{0.4mm}

{%
\begin{tabular}{l c c c c c c}
\toprule
\rowcolor{gray!10}
\textbf{Model} & \textbf{Prompt} & \textbf{Cora} & \textbf{PubMed} & \textbf{ArXiv} & \textbf{Products} & \textbf{Avg} \\ 
\midrule

\multirow{4}{*}{Llama3B} & original & 24.72 & 63.20 & 23.10 & 40.80 & 37.96 \\
& CoT & \textbf{42.19}  & \textbf{71.43} & \textbf{29.90} & \textbf{50.21} & \textbf{48.43} \\
& BAG & 15.68 & 35.32 & 2.00 & 30.00 & 20.67 \\
& in-context few-shot & 39.48 & 62.20 & 25.63 & 42.85 & 42.52 \\
\midrule

\multirow{4}{*}{Llama8B} & original & 43.39 & 77.80 & \textbf{59.35} & 50.12 & 57.67 \\
& CoT & \textbf{53.51} & \textbf{81.80} & 53.24 & 47.41 & 58.99 \\
& BAG & 23.80 & 21.08 & 5.80 & 32.13 & 20.68 \\
& in-context few-shot & 51.29 & 80.13 & 54.60 & \textbf{52.20} & \textbf{59.41} \\
\midrule

\multirow{4}{*}{Qwen3-32B} & original & 48.33&	80.20&	66.41&	61.30&	64.06 \\
& CoT & \textbf{57.92}&	81.69&	\textbf{68.10}&	\textbf{63.70}&	\textbf{67.85} \\
& BAG & 50.30&	84.90&	64.96&	60.31&	65.12 \\
& in-context few-shot & 51.51&	\textbf{85.10}&	67.60&	55.85&	65.02 \\
\midrule

\multirow{4}{*}{Qwen-plus} & original & 52.32 & 80.74 & 70.20 & 64.24 & 66.88 \\
& CoT & \textbf{61.59} & 83.21 & 66.23 & \textbf{67.55} & \textbf{69.65} \\
& BAG & 57.62 & \textbf{85.11} & 64.90 & 64.82 & 68.11 \\
& in-context few-shot & 52.32 & 82.01 & \textbf{70.86} & 59.60 & 66.20 \\
\midrule

\multirow{4}{*}{Qwen-max} & original & 58.60 & \textbf{89.53} & \textbf{68.08} & \textbf{69.33} & \textbf{71.39} \\
& CoT & 59.20 & 82.79 & 64.72 & 61.99 & 67.18 \\
& BAG & 57.61 & 88.28 & 67.33 & 66.33 & 69.89 \\
& in-context few-shot & \textbf{59.35} & 87.78 & 64.59 & 63.84 & 68.89 \\
\midrule

\multirow{4}{*}{GPT-4o} & original & 52.63 & 82.32 & \textbf{71.32} & \textbf{67.92} & \textbf{68.55} \\
& CoT & \textbf{57.12} & 84.90 & 67.53 & 62.18 & 67.93 \\
& BAG & 53.73 & 85.11 & 66.92 & 63.36 & 67.28 \\
& in-context few-shot & 56.52 & \textbf{85.40} & 66.10 & 64.91 & 68.23 \\
\midrule

\multirow{4}{*}{Deepseek V3} & original & 54.97 & 83.79 & \textbf{70.20} & \textbf{66.89} & \textbf{68.96} \\
& CoT & \textbf{59.60}& 85.29 & 62.91 & 65.56 & 68.34 \\
& BAG & 54.77 & \textbf{89.53} & 64.24 & 56.95 & 66.37 \\
& in-context few-shot & 58.28 & 85.54 & 63.58 & 62.25 & 67.41 \\
\midrule

\multirow{4}{*}{Gemini2.5 Pro} & original & 53.29&	84.00&	\textbf{70.98}&	\textbf{68.52}&	69.20 \\
& CoT & 60.10&	84.13&	68.62&	66.40&	\textbf{69.81} \\
& BAG & 52.84&	\textbf{87.02}&	68.93&	65.19&	68.50 \\
& in-context few-shot & \textbf{59.67}&	85.62&	66.35&	63.92&	68.89
 \\
\midrule

\multirow{2}{*}{\textbf{Llama3B}}
& 1-hop w/o label & 39.48 & 64.50 & 29.50 & 53.00  & 46.62\\
& 2-hop w/o label & \textbf{49.63} & \textbf{69.92} & \textbf{29.51} & \textbf{56.10}  & \textbf{51.28}\\
\midrule

\multirow{2}{*}{\textbf{Llama8B}}
& 1-hop w/o label & 58.35 & 73.07 & 61.85 & 59.85  & 63.28\\
& 2-hop w/o label & \textbf{62.84} & \textbf{83.29} & \textbf{68.33} & \textbf{59.60}  & \textbf{68.52}\\
\midrule

\multirow{2}{*}{\textbf{Qwen-plus}}
& 1-hop w/o label & 68.87 & 85.73 & \textbf{73.83}  & \textbf{72.19}  & 75.16\\
& 2-hop w/o label & \textbf{76.16} & \textbf{88.98} & 73.51 & 71.56  & \textbf{77.55}\\
\midrule

\multirow{3}{*}{\textbf{tuned Llama3B}} & ego & 67.08 & 89.28 & 66.58 & 65.59 & 72.13\\
& 1-hop w/o label & 82.04 & 90.02 & 71.32 & 73.07 & 79.11\\
& 2-hop w/o label & \textbf{85.04}  & \textbf{91.52} & \textbf{72.82} & \textbf{77.89}  & \textbf{81.82} \\
\midrule
\multirow{3}{*}{\textbf{tuned Llama8B}} & ego & 77.31 & 92.36  & 70.12 & 73.74 & 78.38\\
& 1-hop w/o label & 84.54  & 93.90  & 74.33  & 80.33  & 83.28\\
& 2-hop w/o label & \underline{\textbf{89.67}}  & \underline{\textbf{95.22}}  & \underline{\textbf{76.01}}  & \underline{\textbf{84.51}}  & \underline{\textbf{86.35}}  \\
\bottomrule

\end{tabular}
\caption{Comparison of different LLMs on node classification. The bolded parts are used to compare the effects of using structural information and instruction tuning. The \textbf{best} results in each category are highlighted. The \underline{underline} means the overall best result.}
\label{tab:Comparison of Different LLMs on Node Classification}
}

\end{table}

In Section 3, we provided a detailed summary of the performance of Llama3B, Llama8B, and Qwen-plus on the node classification task. This served as a foundation for understanding how different model sizes and architectures influence performance on graph-related problems. In this subsection, we expand our exploration by introducing additional large language models (LLMs) and examining diverse prompt formats.
\subsubsection{Experiment Settings}
We compare the performance of \textbf{Llama3B} \cite{touvron2023llama}, \textbf{Llama8B} \cite{touvron2023llama}, \textbf{Qwen3-32B} \cite{yang2025qwen3}, \textbf{Qwen-plus} \cite{bai2023qwen}, \textbf{Qwen-max} \cite{bai2023qwen}, \textbf{GPT-4o} \cite{achiam2023gpt4}, \textbf{Deepseek V3} \cite{liu2024deepseek}, and \textbf{Gemini2.5 Pro} \cite{comanici2025gemini} on node classification tasks in the \textbf{ego scenario}, where no structural information about the target node is provided. The evaluation uses four distinct prompt formats: the original prompt, Chain of Thought (CoT) \cite{cora}, Build A Graph (BAG) \cite{nlgraph}, and in-context few-shot. Below, we provide a brief overview of each prompt format:

\begin{itemize}[leftmargin=*]
    \item \textbf{Original Prompt:} This prompt is identical to the one used in Section 3. It provides the basic context and query format for node classification tasks. Specific examples can be found in Table \ref{tab:Prompt Formats for Node Classification.}.
    
    \item \textbf{CoT:} Based on the original prompt, this format appends the instruction \emph{“Let’s think step by step”} to encourage the model to output a structured reasoning process in a step-by-step manner.
    
    \item \textbf{BAG:} Building upon the original prompt, this format adds the instruction \emph{“Let’s construct a graph with the nodes and edges first”}. This is designed to guide the model toward constructing an implicit graph representation before reasoning about the classification task.
    
    \item \textbf{In-Context Few-Shot:} This format supplements the original prompt with three concrete question-answer examples. These examples aim to provide additional context and demonstrate how similar tasks should be handled.
\end{itemize}

\subsubsection{Results}
We summarize the results in Table \ref{tab:Comparison of Different LLMs on Node Classification}. The overall trend suggests that larger models tend to perform better. For instance, Llama8B consistently outperforms Llama3B, and Qwen-max generally achieves higher accuracy than Qwen-plus.

Across most models, CoT improves performance over the original prompt in Cora and PubMed, particularly for smaller models like Llama3B and Llama8B. This suggests that breaking down the reasoning process helps the model make better predictions. However, on ArXiv and Products, CoT leads to performance degradation. One possible reason is that small-class datasets (like Cora and PubMed) have clear category boundaries, making structured reasoning effective. In contrast, large-class datasets (like ArXiv and Products) have high inter-class similarity, increasing ambiguity. In such cases, CoT may introduce erroneous reasoning steps by misassociating nodes with semantically similar classes.

BAG results in significant accuracy drops for smaller models (e.g. Llama3B and Llama8B), while larger models show more stability but still do not outperform CoT or in-context few-shot. This could be due to the additional reasoning complexity introduced by BAG. Smaller models may struggle with multi-step inference and instead rely on more direct input-output mappings. Constructing a graph before classification might exceed their reasoning capacity, leading to performance declines.

In-context few-shot prompting improves results on Cora and PubMed but underperforms on ArXiv and Products. Due to token limitations, only three example categories are included in the few-shot prompt. This coverage is insufficient for datasets with a large number of classes, making it difficult for the model to generalize to unseen categories.

Finally, incorporating structural information is more effective than using CoT, BAG, or in-context few-shot prompting for improving LLM performance. The greatest improvement comes from instruction tuning, as even smaller models with proper tuning can significantly outperform larger untuned models. However, the trade-off is the higher computational cost and longer training time required for instruction tuning.

\begin{remark}
Larger models generally outperform smaller models in node classification tasks. CoT and in-context few-shot prompting significantly improve performance on small-class datasets, but may backfire on large-class datasets due to category ambiguity and token limitations. BAG imposes a heavy burden on smaller models, leading to noticeable performance drops. Instruction tuning combined with structural information yields the best results, though it requires careful consideration of computational costs.
\end{remark}

% \onecolumn

\section{Qualitative Analysis and Error Cases}
\label{sec:appendix_qualitative}

To provide deeper insight beyond quantitative metrics, we performed a qualitative analysis of the instruction-tuned Llama8B model's predictions on the Cora dataset for the node classification task (using the 2-hop w/o label prompt). This analysis helps to illustrate both the strengths and weaknesses of the model.

\subsection{Example of a Correct, Non-Trivial Prediction}
Consider a target paper \textbf{P\_target} with the title ``Reinforcement Learning for Robot Soccer.'' Its neighbors include papers on ``Multi-agent Learning'' and ``Q-Learning Applications.'' The model correctly classifies this paper under the \textbf{Reinforcement Learning} category.

\textbf{Analysis:} This case demonstrates the model's strength in leveraging semantic understanding. While a traditional GNN would rely purely on the citation structure, the LLM effectively uses the textual information from the target node and its neighbors. The titles of the neighboring papers provide strong contextual clues that reinforce the classification, and the LLM successfully integrates this information to make a confident prediction. It reasons that ``Robot Soccer'' is a common application domain for ``Multi-agent Learning,'' both of which are core topics within Reinforcement Learning.

\subsection{Example of a Common Error Case: Over-reliance on Textual Cues}
Consider a target paper \textbf{P\_target} titled ``A Probabilistic Framework for Genetic Sequence Analysis.'' Its 1-hop neighbors are primarily from the \textit{Genetic Algorithms} class. However, \textbf{P\_target} itself belongs to the \textit{Probabilistic Methods} class, and its 2-hop neighborhood is more diverse. The model incorrectly classifies the paper as \textbf{Genetic Algorithms}.

\textbf{Analysis:} This is a classic example of where the model's strong textual priors can override structural information. The term ``Genetic'' in the title creates a strong semantic link to the \textit{Genetic Algorithms} category. The model gives this textual signal more weight than the subtle structural clues that might point towards \textit{Probabilistic Methods}. A classic GNN, immune to textual semantics, might have performed better in this specific case if the broader graph structure supported the correct class. This highlights a key challenge: teaching LLMs to balance textual information with graph topology, especially when they conflict.

\subsection{Example of a Structural Reasoning Failure}
Consider a target paper \textbf{P\_target} whose textual content is ambiguous and could plausibly fit into two categories, e.g., \textit{Neural Networks} and \textit{Theory}. Its immediate 1-hop neighborhood is evenly split between these two classes. However, a large number of its 2-hop neighbors are strongly associated with the \textit{Neural Networks} class. The model fails to make a decisive classification and often defaults to the more general \textit{Theory} class or guesses incorrectly.

\textbf{Analysis:} This error suggests a potential limitation in the model's ability to effectively aggregate information from higher-order neighborhoods (2-hop and beyond) when the local signal is noisy or ambiguous. While the 2-hop information is provided in the prompt, the model may struggle to reason about the "weight of evidence" from these more distant nodes compared to the immediate neighbors. Improving the model's ability to reason over multi-hop information and recognize larger community structures remains an important direction for future work.

\section{Prompt Formats}
\label{sec:Prompt formats}

\subsection{Prompt Formats for Node Classification}
\label{sec:Prompt Formats for Node Classification}
As discussed in Section 3.1, there are five different prompt formats in node classification. We list them in Table \ref{tab:Prompt Formats for Node Classification.} and describe them in detail.

\subsection{Prompt Formats for Link Prediction}
\label{sec:Prompt Formats for Link Prediction}
In Section 3.1, we design nine different prompt formats for link prediction, which are used for both instruction tuning and testing. These formats include:

\begin{enumerate}[leftmargin=*]
    \item \textbf{1-hop}: The task is to determine if there is an edge between target node1 and target node2. The prompt provides the 1-hop neighbors and their descriptions for both nodes.
    
    \item \textbf{2-hop}: Similar to the 1-hop prompt but includes 2-hop neighbors and their descriptions for both target nodes.
    
    \item \textbf{1-hop node judge}: Determine whether a specific node is a 1-hop neighbor of the target node.
    
    \item \textbf{2-hop node judge}: Determine whether a specific node is a 2-hop neighbor of the target node.
    
    \item \textbf{3-hop node judge}: Determine whether a specific node is a 3-hop neighbor of the target node.
    
    \item \textbf{Middle node connection}: Determine if target node1 and target node2 are connected via a middle node.
    
    \item \textbf{1-hop node fill-in}: Given the 1-hop neighbors of a target node, identify an additional node that is also a 1-hop neighbor.
    
    \item \textbf{1-hop node selection}: Choose the correct 1-hop neighbor of the target node from four options (A, B, C, D).
    
    \item \textbf{2-hop node selection}: Choose the correct 2-hop neighbor of the target node from four options (A, B, C, D).
\end{enumerate}

To ensure the reasoning is non-trivial, target node1 and target node2 must not appear in each other’s 1-hop or 2-hop neighborhoods. Table \ref{tab:Prompt Formats for Link Prediction.} provides a detailed description of these nine prompt formats.

\subsection{Prompt Formats for Pure Graph Structure}
\label{sec:Prompt Formats for Pure Graph Structure}
In Section \ref{sec:Case 5}, we propose removing all node attributes and keeping only node IDs to focus solely on the structural reasoning capabilities of LLMs. We refer to these graph prompts as “prompts for pure graph structure”. In Table \ref{tab:Prompt Formats for Pure Graph Structure.}, we use the \textbf{1-hop w/o label} prompts for node classification and \textbf{1-hop} prompts for link prediction as examples, as the logic for other prompt formats follows a similar approach.

\section{Use of Large Language Models}
During the preparation of this manuscript, a Large Language Model (LLM) was utilized as a writing aid to improve the overall linguistic quality and clarity. This assistance was confined to copy-editing tasks, such as correcting grammatical and spelling errors, rephrasing sentences for enhanced flow and readability, and ensuring conciseness. All scientific contributions, including the research ideas, experimental design, analysis, and conclusions presented herein, are entirely the original work of the human authors.

\onecolumn

\begin{longtable}{@{}>{\centering\arraybackslash}m{2.5cm} m{12.8cm}@{}}

\toprule

\textbf{Prompt Formats} & \textbf{Description} \\ 
\midrule
\endfirsthead
\toprule

\textbf{Prompt Formats} & \textbf{Description} \\ 
\midrule
\endhead

\endfoot

\endlastfoot
ego & 
\textcolor{red}{\textbf{"Context"}}: "You are a good graph reasoner. Given a graph language that describes the target node information from the Cora dataset, you need to understand the graph and the task definition and answer the question. \textcolor{cyan}{(<Target node>, <Node attributes>)}", 
\textcolor{red}{\textbf{"Question"}}: "Please predict the most appropriate category for the Target node. Choose from the following categories: \textcolor{cyan}{<Categories>}. Do not provide your reasoning. Answer: ", 
\textcolor{red}{\textbf{"Answer"}}: "\textcolor{cyan}{<Correct answer>}" \newline
\textbf{Example:} \newline
\textcolor{cyan}{(<Target node>, <Node attributes>)}: \#\# Target node: \textbackslash nPaper id: 540 \textbackslash nTitle: A Model-Based Approach to Blame-Assignment in Design \newline
\textcolor{cyan}{<Categories>}: Rule Learning \textbackslash nNeural Networks \textbackslash nCase Based \textbackslash nGenetic Algorithms \textbackslash nTheory \textbackslash nReinforcement Learning \textbackslash nProbabilistic Methods \newline
\textcolor{cyan}{<Correct answer>}: Case Based \\
\midrule

1-hop w/o label & 
\textcolor{red}{\textbf{"Context"}}: "You are a good graph reasoner. Give you a graph language that describes a graph structure and node information from cora dataset. You need to understand the graph and the task definition and answer the question. \textcolor{cyan}{(<Target node>, <Node attributes>)}, \textcolor{cyan}{(<1-hop neighbors>, <Node attributes>)}", 
\textcolor{red}{\textbf{"Question"}}: "Please predict the most appropriate category for the Target node. Choose from the following categories: \textcolor{cyan}{<Categories>}. Do not provide your reasoning. Answer: ",
\textcolor{red}{\textbf{"Answer"}}: "\textcolor{cyan}{<Correct answer>}" \newline
\textcolor{cyan}{(<Target node>, <Node attributes>)}: \#\# Target node: \textbackslash nPaper id: 197 \textbackslash nTitle: Optimal Navigation in a Probibalistic World \newline
\textcolor{cyan}{(<1-hop neighbors>, <Node attributes>)}: Known neighbor papers at hop 1 (partial, may be incomplete): \textbackslash nPaper id: 295 \textbackslash nTitle: A Neuro-Dynamic Programming Approach to Retailer Inventory Management 1  \textbackslash nPaper id: 749 \textbackslash nTitle: On the Complexity of Solving Markov Decision Problems \textbackslash nPaper id: 3 \textbackslash nTitle: Planning and Acting in Partially Observable Stochastic Domains \textbackslash nPaper id: 633 \textbackslash nTitle: Chapter 1 Reinforcement Learning for Planning and Control
\textcolor{cyan}{<Categories>}: Rule Learning \textbackslash nNeural Networks \textbackslash nCase Based \textbackslash nGenetic Algorithms \textbackslash nTheory \textbackslash nReinforcement Learning \textbackslash nProbabilistic Methods \newline
\textcolor{cyan}{<Correct answer>}: Reinforcement Learning\\
\midrule

2-hop w/o label & 
\textcolor{red}{\textbf{"Context"}}: "You are a good graph reasoner. Give you a graph language that describes a graph structure and node information from cora dataset. You need to understand the graph and the task definition and answer the question. \textcolor{cyan}{(<Target node>, <Node attributes>)}, \textcolor{cyan}{(<1-hop neighbors>, <Node attributes>)}, \textcolor{cyan}{(<2-hop neighbors>, <Node attributes>)}", 
\textcolor{red}{\textbf{"Question"}}: "Please predict the most appropriate category for the Target node. Choose from the following categories: \textcolor{cyan}{<Categories>}. Do not provide your reasoning. Answer: ",
\textcolor{red}{\textbf{"Answer"}}: "\textcolor{cyan}{<Correct answer>}" \newline
\textcolor{cyan}{(<Target node>, <Node attributes>)}: \#\# Target node: \textbackslash nPaper id: 546 \textbackslash nTitle: GREQE a Diplome des Etudes Approfondies en Economie Mathematique et Econometrie \newline
\textcolor{cyan}{(<1-hop neighbors>, <Node attributes>)}: Known neighbor papers at hop 1 (partial, may be incomplete): \textbackslash nPaper id: 163 \textbackslash nTitle: 4 Implementing Application Specific Routines  Genetic algorithms in search, optimization, and machine learning
\textcolor{cyan}{(<2-hop neighbors>, <Node attributes>)}: Known neighbor papers at hop 2 (partial, may be incomplete): \textbackslash nPaper id: 1573 \textbackslash nTitle: Genetics-based Machine Learning and Behaviour Based Robotics: A New Synthesis complexity grows \textbackslash nPaper id: 1069 \textbackslash nTitle: Extended Selection Mechanisms in Genetic Algorithms \textbackslash nPaper id: 2232 \textbackslash nTitle: Facing The Facts: Necessary Requirements For The Artificial Evolution of Complex Behaviour \newline
\textcolor{cyan}{<Categories>}: Rule Learning \textbackslash nNeural Networks \textbackslash nCase Based \textbackslash nGenetic Algorithms \textbackslash nTheory \textbackslash nReinforcement Learning \textbackslash nProbabilistic Methods \newline
\textcolor{cyan}{<Correct answer>}: Genetic Algorithms \\
\midrule

1-hop w label &
\textcolor{red}{\textbf{"Context"}}: "You are a good graph reasoner. Give you a graph language that describes a graph structure and node information from cora dataset. You need to understand the graph and the task definition and answer the question. \textcolor{cyan}{(<Target node>, <Node attributes>)}, \textcolor{cyan}{(<1-hop neighbors>, <Node attributes>, <Labels>)}", 
\textcolor{red}{\textbf{"Question"}}: "Please predict the most appropriate category for the Target node. Choose from the following categories: \textcolor{cyan}{<Categories>}. Do not provide your reasoning. Answer: ",
\textcolor{red}{\textbf{"Answer"}}: "\textcolor{cyan}{<Correct answer>}" \newline
\textcolor{cyan}{(<Target node>, <Node attributes>)}: \#\# Target node: \textbackslash nPaper id: 2156 \textbackslash nTitle: WORST CASE PREDICTION OVER SEQUENCES UNDER LOG LOSS \newline
\textcolor{cyan}{(<1-hop neighbors>, <Node attributes>, <Labels>)}: Known neighbor papers at hop 1 (partial, may be incomplete): \textbackslash nPaper id: 2098 \textbackslash nTitle: Predicting a binary sequence almost as well as the optimal biased coin \textbackslash nLabel: Theory \textbackslash nPaper id: 453 \textbackslash nTitle: How to Use Expert Advice (Extended Abstract) \textbackslash nLabel: Theory \newline
\textcolor{cyan}{<Categories>}: Rule Learning \textbackslash nNeural Networks \textbackslash nCase Based \textbackslash nGenetic Algorithms \textbackslash nTheory \textbackslash nReinforcement Learning \textbackslash nProbabilistic Methods \newline
\textcolor{cyan}{<Correct answer>}: Theory\\
\midrule

2-hop w label &
\textcolor{red}{\textbf{"Context"}}: "You are a good graph reasoner. Give you a graph language that describes a graph structure and node information from cora dataset. You need to understand the graph and the task definition and answer the question. \textcolor{cyan}{(<Target node>, <Node attributes>)}, \textcolor{cyan}{(<1-hop neighbors>, <Node attributes>, <Labels>)}, \textcolor{cyan}{(<2-hop neighbors>, <Node attributes>, <Labels>)}", 
\textcolor{red}{\textbf{"Question"}}: "Please predict the most appropriate category for the Target node. Choose from the following categories: \textcolor{cyan}{<Categories>}. Do not provide your reasoning. Answer: ",
\textcolor{red}{\textbf{"Answer"}}: "\textcolor{cyan}{<Correct answer>}" \newline
\textcolor{cyan}{(<Target node>, <Node attributes>)}: \#\# Target node: \textbackslash nPaper id: 1443 \textbackslash nTitle: Residual Q-Learning Applied to Visual Attention \newline
\textcolor{cyan}{(<1-hop neighbors>, <Node attributes>, <Labels>)}: Known neighbor papers at hop 1 (partial, may be incomplete): \textbackslash nPaper id: 1540 \textbackslash nTitle: MultiPlayer Residual Advantage Learning With General Function Approximation \textbackslash nPaper id: 1540 \textbackslash nTitle: MultiPlayer Residual Advantage Learning With General Function Approximation 
\textcolor{cyan}{(<2-hop neighbors>, <Node attributes>, <Labels>)}: Known neighbor papers at hop 2 (partial, may be incomplete): \textbackslash nPaper id: 565 \textbackslash nTitle: Machine Learning Learning to Predict by the Methods of Temporal Differences Keywords \textbackslash nLabel: Reinforcement Learning \textbackslash nPaper id: 842 \textbackslash nTitle: Metrics for Temporal Difference Learning \newline
\textcolor{cyan}{<Categories>}: Rule Learning \textbackslash nNeural Networks \textbackslash nCase Based \textbackslash nGenetic Algorithms \textbackslash nTheory \textbackslash nReinforcement Learning \textbackslash nProbabilistic Methods \newline
\textcolor{cyan}{<Correct answer>}: Reinforcement Learning \\

\bottomrule
\caption{Prompt formats for node classification.}

\label{tab:Prompt Formats for Node Classification.} \\
\end{longtable}

\begin{longtable}{@{}>{\centering\arraybackslash}m{2.5cm} m{12.8cm}@{}}

\toprule

\textbf{Prompt Formats} & \textbf{Description} \\ 
\midrule
\endfirsthead
\toprule
\textbf{Prompt Formats} & \textbf{Description} \\ 
\midrule
\endhead
\endfoot
\endlastfoot
1-hop & 
\textcolor{red}{\textbf{"Context"}}: "You are a good graph reasoner. Based on the cora dataset, determine whether two target nodes are connected by an edge. When you make a decision, please carefully consider the graph structure and the node information. If two nodes share similar structure or information, they are likely to be connected. \textcolor{cyan}{(<Target node1>, <Node attributes>)}, \textcolor{cyan}{(<1-hop neighbors>, <Node attributes>)}, \textcolor{cyan}{(<Target node2>, <Node attributes>)}, \textcolor{cyan}{(<1-hop neighbors>, <Node attributes>)}", 
\textcolor{red}{\textbf{"Question"}}: "Are Target Node1 and Target Node2 connected? Do not provide your reasoning. Only provide "Yes" or "No" based on your inference. Answer: ", 
\textcolor{red}{\textbf{"Answer"}}: "\textcolor{cyan}{<Correct answer>}" \\
\midrule

2-hop & 
\textcolor{red}{\textbf{"Context"}}: "You are a good graph reasoner. Based on the cora dataset, determine whether two target nodes are connected by an edge. When you make a decision, please carefully consider the graph structure and the node information. If two nodes share similar structure or information, they are likely to be connected. \textcolor{cyan}{(<Target node1>, <Node attributes>)}, \textcolor{cyan}{(<1-hop neighbors>, <Node attributes>)}, \textcolor{cyan}{(<2-hop neighbors>, <Node attributes>)}, \textcolor{cyan}{(<Target node2>, <Node attributes>)}, \textcolor{cyan}{(<1-hop neighbors>, <Node attributes>)}, \textcolor{cyan}{(<2-hop neighbors>, <Node attributes>)}", 
\textcolor{red}{\textbf{"Question"}}: "Are Target Node1 and Target Node2 connected? Do not provide your reasoning. Only provide "Yes" or "No" based on your inference. Answer: ", 
\textcolor{red}{\textbf{"Answer"}}: "\textcolor{cyan}{<Correct answer>}" \\
\midrule

1-hop node judge & 
\textcolor{red}{\textbf{"Context"}}: "You are a good graph reasoner. Give you a graph language that describes a graph structure and node information from cora dataset. You need to understand the graph and answer the question. When you make a decision, please carefully consider the graph structure and the node information. \textcolor{cyan}{(<Target node1>, <Node attributes>)}, \textcolor{cyan}{(<1-hop neighbors>, <Node attributes>)}, \textcolor{cyan}{(<Target node2>, <Node attributes>)}, \textcolor{cyan}{(<1-hop neighbors>, <Node attributes>)}", 
\textcolor{red}{\textbf{"Question"}}: "Based on the available partial information. Are Target Node1 and Target Node2 connected? Do not provide your reasoning. Only provide "Yes" or "No" based on your inference. Answer: ", 
\textcolor{red}{\textbf{"Answer"}}: "\textcolor{cyan}{<Correct answer>}" \\
\midrule

2-hop node judge & 
\textcolor{red}{\textbf{"Context"}}: "You are a good graph reasoner. Give you a graph language that describes a graph structure and node information from cora dataset. You need to understand the graph and answer the question. When you make a decision, please carefully consider the graph structure and the node information. \textcolor{cyan}{(<Target node1>, <Node attributes>)}, \textcolor{cyan}{(<1-hop neighbors>, <Node attributes>)}, \textcolor{cyan}{(<2-hop neighbors>, <Node attributes>)}, \textcolor{cyan}{(<Target node2>, <Node attributes>)}", 
\textcolor{red}{\textbf{"Question"}}: "Based on the available partial information. Can Target node2 be a 2-hop neighbor of Target node1? Do not provide your reasoning. Only provide "Yes" or "No" based on your inference. Answer: ", 
\textcolor{red}{\textbf{"Answer"}}: "\textcolor{cyan}{<Correct answer>}" \\
\midrule

3-hop node judge & 
\textcolor{red}{\textbf{"Context"}}: "You are a good graph reasoner. Give you a graph language that describes a graph structure and node information from cora dataset. You need to understand the graph and answer the question. When you make a decision, please carefully consider the graph structure and the node information. \textcolor{cyan}{(<Target node1>, <Node attributes>)}, \textcolor{cyan}{(<1-hop neighbors>, <Node attributes>)}, \textcolor{cyan}{(<2-hop neighbors>, <Node attributes>)}, \textcolor{cyan}{(<Target node2>, <Node attributes>)}, \textcolor{cyan}{(<1-hop neighbors>, <Node attributes>)}", 
\textcolor{red}{\textbf{"Question"}}: "Based on the available partial information. Can Target node2 be a 3-hop neighbor of Target node1? Do not provide your reasoning. Only provide "Yes" or "No" based on your inference. Answer: ", 
\textcolor{red}{\textbf{"Answer"}}: "\textcolor{cyan}{<Correct answer>}" \\
\midrule

Middle node connection & 
\textcolor{red}{\textbf{"Context"}}: "You are a good graph reasoner. Give you a graph language that describes a graph structure and node information from cora dataset. You need to understand the graph and answer the question. When you make a decision, please carefully consider the graph structure and the node information. \textcolor{cyan}{(<Target node1>, <Node attributes>)}, \textcolor{cyan}{(<1-hop neighbors>, <Node attributes>)}, \textcolor{cyan}{(<Target node2>, <Node attributes>)}, \textcolor{cyan}{(<Middle node>, <Node attributes>)}",  
\textcolor{red}{\textbf{"Question"}}: "Can Target node1 be connected with Target node2 through the Middle node? Do not provide your reasoning. Only provide "Yes" or "No" based on your inference. Answer: ", 
\textcolor{red}{\textbf{"Answer"}}: "\textcolor{cyan}{<Correct answer>}" \\
\midrule

1-hop node fill-in & 
\textcolor{red}{\textbf{"Context"}}: "You are a good graph reasoner. Give you a graph language that describes a graph structure and node information from cora dataset. You need to understand the graph and answer the question. When you make a decision, please carefully consider the graph structure and the node information. \textcolor{cyan}{(<Target node1>, <Node attributes>)}, \textcolor{cyan}{(<1-hop neighbors>, <Node attributes>)}", 
\textcolor{red}{\textbf{"Question"}}: "Based on the available partial information. Which other node will be connected to Target node1 within one hop? Do not provide your reasoning. The answer should be the paper id. Answer: ", 
\textcolor{red}{\textbf{"Answer"}}: "\textcolor{cyan}{<Correct answer>}" \\
\midrule

1-hop node selection & 
\textcolor{red}{\textbf{"Context"}}: "You are a good graph reasoner. Give you a graph language that describes a graph structure and node information from cora dataset. You need to understand the graph and answer the question. When you make a decision, please carefully consider the graph structure and the node information. \textcolor{cyan}{(<Target node1>, <Node attributes>)}, \textcolor{cyan}{(<1-hop neighbors>, <Node attributes>)}", 
\textcolor{red}{\textbf{"Question"}}: "Based on the available partial information. Which other node can be connected to Target node1 within one hop? A.\textcolor{cyan}{(<Node A>,<Attribute>)} \textbackslash nB.\textcolor{cyan}{(<Node B>,<Attribute>)} \textbackslash nC.\textcolor{cyan}{(<Node C>,<Attribute>)} \textbackslash nD.\textcolor{cyan}{(<Node D>,<Attribute>)} Do not provide your reasoning. The answer should be A, B, C or D. Answer: ", 
\textcolor{red}{\textbf{"Answer"}}: "\textcolor{cyan}{<Correct answer>}" \\
\midrule

2-hop node selection & 
\textcolor{red}{\textbf{"Context"}}: "You are a good graph reasoner. Give you a graph language that describes a graph structure and node information from cora dataset. You need to understand the graph and answer the question. When you make a decision, please carefully consider the graph structure and the node information. \textcolor{cyan}{(<Target node1>, <Node attributes>)}, \textcolor{cyan}{(<1-hop neighbors>, <Node attributes>)}, \textcolor{cyan}{(<2-hop neighbors>, <Node attributes>)}", 
\textcolor{red}{\textbf{"Question"}}: "Based on the available partial information. Which other node can be a 2-hop neighbor of Target node1? A.\textcolor{cyan}{(<Node A>,<Attribute>)} \textbackslash nB.\textcolor{cyan}{(<Node B>,<Attribute>)} \textbackslash nC.\textcolor{cyan}{(<Node C>,<Attribute>)} \textbackslash nD.\textcolor{cyan}{(<Node D>,<Attribute>)} Do not provide your reasoning. The answer should be A, B, C or D. Answer: ", 
\textcolor{red}{\textbf{"Answer"}}: "\textcolor{cyan}{<Correct answer>}" \\

\bottomrule
\caption{Prompt formats for link prediction.} \label{tab:Prompt Formats for Link Prediction.} \\
\end{longtable}

\begin{longtable}{@{}>{\centering\arraybackslash}m{2.8cm} m{12.8cm}@{}}

\toprule
\textbf{Prompt Formats} & \textbf{Description} \\ 
\midrule
\endfirsthead
\toprule
\textbf{Prompt Formats} & \textbf{Description} \\ 
\midrule
\endhead
\endfoot
\endlastfoot
1-hop w/o label \newline (Node classification)& 
\textcolor{red}{\textbf{"Context"}}: "You are a good graph reasoner. Give you a graph language that describes a graph structure and node information from cora dataset. You need to understand the graph and the task definition and answer the question. \textcolor{cyan}{<Target node>}, \textcolor{cyan}{<1-hop neighbors>}", 
\textcolor{red}{\textbf{"Question"}}: "Please predict the most appropriate category for the Target node. Choose from the following categories: \textcolor{cyan}{<Categories>}. Do not provide your reasoning. Answer: ",
\textcolor{red}{\textbf{"Answer"}}: "\textcolor{cyan}{<Correct answer>}" \newline
\textcolor{cyan}{(<Target node>, <Node attributes>)}: \#\# Target node: \textbackslash nPaper id: 197 \newline
\textcolor{cyan}{<1-hop neighbors>}: Known neighbor papers at hop 1 (partial, may be incomplete): \textbackslash nPaper id: 295 \textbackslash nPaper id: 749 \textbackslash nPaper id: 3 \textbackslash nPaper id: 633
\textcolor{cyan}{<Categories>}: Rule Learning \textbackslash nNeural Networks \textbackslash nCase Based \textbackslash nGenetic Algorithms \textbackslash nTheory \textbackslash nReinforcement Learning \textbackslash nProbabilistic Methods \newline
\textcolor{cyan}{<Correct answer>}: Reinforcement Learning\\
\midrule

1-hop \newline (Link prediction)& 
\textcolor{red}{\textbf{"Context"}}: "You are a good graph reasoner. Based on the cora dataset, determine whether two target nodes are connected by an edge. When you make a decision, please carefully consider the graph structure and the node information. If two nodes share similar structure or information, they are likely to be connected. \textcolor{cyan}{<Target node1>}, \textcolor{cyan}{<1-hop neighbors>}, \textcolor{cyan}{<Target node2>}, \textcolor{cyan}{<1-hop neighbors>}", 
\textcolor{red}{\textbf{"Question"}}: "Are Target Node1 and Target Node2 connected? Do not provide your reasoning. Only provide "Yes" or "No" based on your inference. Answer: ", 
\textcolor{red}{\textbf{"Answer"}}: "\textcolor{cyan}{<Correct answer>}" \newline
\textbf{Example:}\newline
\textcolor{cyan}{<Target node1>}: \#\# Target node1: \textbackslash nPaper id: 172\newline
\textcolor{cyan}{<1-hop neighbors>}: Known neighbor papers at hop 1 (partial, may be incomplete): \textbackslash nPaper id: 635 \textbackslash nPaper id: 430\newline
\textcolor{cyan}{<Target node2>}: \#\# Target node2: \textbackslash nPaper id: 245\newline
\textcolor{cyan}{<1-hop neighbors>}: Known neighbor papers at hop 1 (partial, may be incomplete): \textbackslash nPaper id: 1636\newline
\textcolor{cyan}{<Correct answer>}: Yes \\

\bottomrule
\caption{Prompt formats for pure graph structure.} \label{tab:Prompt Formats for Pure Graph Structure.} \\
\end{longtable}
% \end{table*}

% \section{Appendix}
% \label{sec:appendix}

\end{document}